\documentclass[lettersize,journal]{IEEEtran}
\usepackage{amsmath,amsfonts}
\usepackage{algorithmic}
\usepackage{algorithm}
\usepackage{array}
\usepackage[caption=false,font=normalsize,labelfont=sf,textfont=sf]{subfig}
\usepackage{textcomp}
\usepackage{stfloats}
\usepackage{url}
\usepackage{verbatim}
\usepackage{graphicx}
\usepackage{cite}
\newtheorem{thm}{Theorem}[section]
\newtheorem{lem}[thm]{Lemma}

\newtheorem{defn}{Definition}[section]
\newtheorem{remark}{Remark}[section]

\newcommand{\proof}{\noindent {\bf Proof.}~}
\newcommand{\qed}{\hfill \square}

\DeclareMathOperator*{\diag}{diag}

\DeclareMathOperator{\Ima}{Im}

\newcommand{\bR}{\mathbb{R}}

\newcommand{\me}{\mathrm{e}}

\begin{document}

\title{Random orthogonal additive filters: a solution to
the vanishing/exploding gradient of deep neural networks}

\author{Andrea Ceni
\thanks{The author is with the Department of Computer Science, University of Pisa, Largo Bruno Pontecorvo, 3 - 56127, IT (e-mail:
andrea.ceni@di.unipi.it), and the Department of Mathematics, University of Exeter, Exeter EX4 4QF, UK (e-mail:
ac860@exeter.ac.uk).
}
\thanks{Manuscript received September ??, 2022; revised September ??, 2022.}}



\maketitle

\begin{abstract}
Since the recognition in the early nineties of the vanishing/exploding (V/E) gradient issue plaguing the training of neural networks (NNs), significant efforts have been exerted to overcome this obstacle. 
However, a clear solution to the V/E issue remained elusive so far.
In this manuscript a new architecture of NN is proposed, designed to mathematically prevent the V/E issue to occur. 
The pursuit of approximate dynamical isometry, i.e. parameter configurations where the singular values of the input-output Jacobian are tightly distributed around 1, leads to the derivation of a NN's architecture that shares common traits with the popular Residual Network model. 
Instead of skipping connections between layers, the idea is to filter the previous activations orthogonally and add them to the nonlinear activations of the next layer, realising a convex combination between them.
Remarkably, the impossibility for the gradient updates to either vanish or explode is demonstrated with analytical bounds that hold even in the infinite depth case.
The effectiveness of this method is empirically proved by means of training via backpropagation an extremely deep multilayer perceptron of 50k layers, and an Elman NN to learn long-term dependencies in the input of 10k time steps in the past.
Compared with other architectures specifically devised to deal with the V/E problem, e.g. LSTMs for recurrent NNs, the proposed model is way simpler yet more effective.
Surprisingly, a single layer vanilla RNN can be enhanced to reach state of the art performance, while converging super fast; for instance on the psMNIST task, it is possible to get test accuracy of over $94 \%$ in the first epoch, and over $98 \%$ after just 10 epochs.
\end{abstract}


\section{Introduction}
\label{sec:introduction}
\IEEEPARstart{A}{rtificial} deep neural networks (NNs) are computational models that have gathered massive interest in the last decade \cite{lecun2015deep}. Usually NNs are composed of many simple units (neurons) organised in multiple layers which nonlinearly interact to each other via trainable parameters, also called weights.
Roughly speaking, training a NN to solve a task means to adapt its parameters in order to 
fit an unknown function mapping between raw input and the output labels forming the data points from which to generalise.

Although several algorithms exist for training NN models, the backpropagation algorithm \cite{rumelhart1986learning, rumelhart1985learning} with stochastic gradient descent (or variants thereof \cite{ruder2016overview}) has been established as a standard for supervised learning.
The backpropagation is a clever procedure to assess how sensitive the output error is w.r.t. a given parameter of the model; thus, each parameter is modified proportionally to this measure of sensitivity. 
The computation involves the sequential product of many matrices whose resulting norm, similarly to the product of many real numbers, can shrink to vanish or expand to infinity exponentially with depth.
This problem has been known in the literature as the vanishing/exploding (V/E) gradient issue.
The V/E issue has been firstly recognised in the context of recurrent neural networks (RNNs) in the nineties \cite{hochreiter1991untersuchungen,bengio1994learning}, and then, after the advent of increasingly deep architectures, it became evident also in feedforward neural networks (FNNs) \cite{glorot2010understanding}.

\subsection{Related works}

Extensive works have been made to combat the V/E gradient issue. Some of the main strategies are reported in this section.

Based on the particular choice of activation function, it has been proved that some weights initialisation, known as xavier-init \cite{glorot2010understanding} and kaimin-init \cite{he2015delving}, generate favourable statistics in the neuronal activations that alleviate the issue.

The V/E dynamics have been recognised to be stronger in “squashing" activation functions, e.g. tanh, while been gentler on hard-nonlinear activation functions, e.g. ReLUs. Moreover, true zero neuronal activations induced by ReLUs promotes sparsity into the network which in turn has been recognised to be beneficial for backpropagating errors, as long as some neural paths remain actives \cite{glorot2011deep}. Nevertheless the issue, although mitigated, is still present with ReLUs. 

Moreover, a mean-field theory analysis on linear FNNs revealed the effectiveness of orthogonal weights initialisation schemes \cite{saxe2013exact}, and, provided that an NN operates close to the so-called \emph{edge of chaos} \cite{bertschinger2004real, legenstein2007edge}, on nonlinear FNNs as well. In fact, the V/E gradient issue can be solved by promoting the NN to operate in an approximate \emph{dynamical isometry} regime, i.e. a configuration where the distribution of the singular values of the NN's input-output Jacobian remains circumscribed around 1 without spreading too far from it. This condition would ensure that signals can flow back and forth along the layered architecture. However, containing the learning process into an approximate dynamical isometry regime appears to be hard in general FNNs, due to the nonlinearities between layers.
In \cite{pennington2017resurrecting} a theoretical analysis about orthogonal initialisations led to the conclusion that ReLU networks are incapable of dynamical isometry, while sigmoid and tanh networks can. Orthogonal initialisation schemes \cite{mishkin2015all,xiao2018dynamical}, although not directly solving the V/E problem, have been successfully implemented to train extremely deep NNs.

Additionally, gradient clipping procedures \cite{pascanu2013difficulty} have been extensively used to maintain the gradients at a controlled magnitude, while not addressing the vanishing problem.
Besides, normalising the incoming input from a layer to the consecutive one, in order to avoid the saturating tails, has been shown to accelerate training while improving the performance \cite{ioffe2015batch}.
However, these are passive strategies that do not aim to prevent the V/E gradient issue to occur.

Another breakthrough is represented by Highway Networks \cite{srivastava2015highway}, a feedforward architecture, inspired by the popular LSTM recurrent NN model, provided with trainable gating mechanisms to regulate information flow.
Finally, Residual Networks (ResNets) \cite{he2016deep}, a gateless version of Highway Networks, have revolutionised the field of computer vision. These architectures provide shortcut paths, directly connecting non-consecutive layers, de facto realising ensembles of relatively shallow networks \cite{veit2016residual}.
Despite the important empirical achievements of ResNets, theoretical results establishing the effective solution of the V/E gradient problem for those architectures are still missing.
In fact, although skip connections have been observed to drive ResNets to the edge of chaos \cite{yang2017mean}, ResNets still heavily rely on batch normalisation for good results with very deep architectures \cite{de2020batch}. 

In practice, it is not rare in the literature to encounter several of the aforementioned techniques alchemically combined together to train deep learning models.\\

RNNs have been especially known for the difficulty to train them to learn long-term dependencies. 
Probably the most popular solution proposed to combat the V/E issue of RNNs is represented by gated recurrent architectures, e.g. as LSTMs \cite{hochreiter1997long}, and GRUs \cite{cho2014properties}. However, despite the increased computational complexity of these models, they still suffer from instabilities when trained on very long sequences.

Similarly to FNNs, orthogonal initialisation of the recurrent weights have been investigated and proved to be effective for RNNs \cite{chen2018dynamical,gilboa2019dynamical}.
Interestingly, it has been showed that initialising the recurrent weights to the identity matrix in a vanilla RNN with ReLU activations may reach performance comparable to LSTMs \cite{le2015simple}.

In the seminal work of \cite{arjovsky2016unitary} the authors go beyond the mere orthogonal initialisation and propose to parametrise the recurrent weights in the space of unitary matrices, thereby ensuring the orthogonality condition during the learning process. Then, in \cite{wisdom2016full} they further improve such model allowing the learning dynamics to access the whole submanifold of the unitary matrices. 
However, these methods come at the expense of considerable computational effort.
After these achievements a burst of works have followed in the footsteps of constraining the recurrent weights to be orthogonal or approximately such \cite{lezcano2019cheap,helfrich2018orthogonal,lezcano2019trivializations,jing2017tunable,vorontsov2017orthogonality} trying to restrain the run time.
Although it is not understood how, and to what extent, these restrictions imposed on recurrent weights impact the learning dynamics, it is widely recognised in the literature that imposing unitary constraints on the recurrent matrix affects its expressive power \cite{kerg2019non,kusupati2019fastgrnn}.

Alternatively to orthogonal constraints, IndRNN \cite{li2018independently} is a recently proposed RNN model where hidden-to-hidden weights are constraint to diagonal matrices nonlinearly stacked in two or more layers as a single recurrent block. V/E gradient is thus prevented bounding the coefficients of those diagonal matrices in a suitable range which depends on the length of the temporal dependencies the IndRNN is aimed to learn.

Another option is to prevent the V/E issue at a fundamental level by constraining the recurrent matrix to be constant, i.e. untrained.
An emblematic example is given by Reservoir Computing machines \cite{lukovsevivcius2009reservoir}. Typically, the recurrent weights are randomly generated and left untouched, tuning them just at a hyperparameter level, e.g. rescaling the spectral radius \cite{lukovsevivcius2012practical}, and only an output readout is trained.
A further case is studied in \cite{rotman2020shuffling}, where the hidden-to-hidden weights are fixed as a specific orthogonal matrix (namely a shifting permutation matrix), while all the other weights are trained.

Other important developments come from considering continuous-time RNNs as ODEs. For instance, AntisymmetricRNN \cite{chang2019antisymmetricrnn} uses skew-symmetric recurrent matrices to ensure stable dynamics. Based on control theory results that ensures global exponential stability, LipschitzRNN \cite{erichson2020lipschitz} extends the pool of recurrent matrices where the learning occurs outside the solely space of skew-symmetric matrices. While another promising architecture is coRNN \cite{rusch2020coupled}, derived from a second-order ODE modeling a network of oscillators. 

Other interesting works include models provided with memory cells, e.g. LMUs \cite{voelker2019legendre} and NRUs \cite{chandar2019towards}, and recurrent models based on skip connections \cite{chang2017dilated,wang2016recurrent,yue2018residual}.

\subsection{Original contributions}

In this paper, a new idea that permits to solve the V/E gradient issue of deep learning models trained via stochastic gradient descent (SGD) methods is proposed. 
Empirically, it is demonstrated how to:
\begin{itemize}
    \item enhance the vanilla FNN model to be trainable for extremely deep architectures (e.g. 50k layers), without relying on auxiliary techniques
    as gradient clipping, batch normalisation, nor skipping connections (see Section \ref{sec:double_moon});
    \item enhance the vanilla RNN model to learn very long-term dependencies, outperforming the great majority of recurrent NN models (among which LSTMs) in benchmark tasks, as the Copying Memory, the Adding Problem, and the Permuted Sequential MNIST (see Section \ref{sec:simulations}).
\end{itemize}
Based on the same intuition, two NN models are proposed: (i) roaFNN, which stands for \emph{random orthogonal additive Feedforward NN}, and (ii) roaRNN, which stands for \emph{random orthogonal additive Recurrent NN}.
Those models result slight variations of, respectively, a multilayer perceptron, and an Elman network.
The simplicity of these models permits a formal mathematical analysis of their gradient update dynamics which even holds for the case of infinite depth. In particular,
both lower and upper bounds are provided for
\begin{itemize}
    \item the maximum singular value of the input-output Jacobian of the roaFNN model (Theorem \ref{thm:main_thm}), demonstrating the impossibility for it to either explode or converge to zero;
    \item the entire set of singular values of the input-output Jacobian of the roaRNN model (Theorem \ref{thm:distribution_singvals}), proving that a roaRNN evolves in an approximate dynamical isometry regime by design.
\end{itemize}
Remarkably, these theoretical results are achieved without any constraint on the parameters of the model.

\section{Approaching dynamical isometry in nonlinear deep FNNs}
\label{sec:feedforward}

\subsection{Background}

A multilayer perceptron (MLP) is described via the following equations:
\begin{align}
\label{eq:v}
& y_{l}  = W_l x_l + b_l \\
\label{eq:FFN}
& x_{l+1}  = \phi( y_{l}), \qquad l=0,1, \ldots , L-1.
\end{align}
When dealing with vectors, it is assumed they are column vectors unless otherwise specified.
That said, $x_l \in \mathbb{R}^{N_l} $ is the vector containing the neuronal activations of the $l$-th layer, and $x_L \in \mathbb{R}^{N_L} $ is the output vector corresponding to the input vector $x_0 \in \mathbb{R}^{N_0}$.
Matrices $W_l \in \mathbb{R}^{N_{l+1} \times N_{l} }$ represent the connection weights from neurons of the $l$-th layer to the $l+1$-th layer, $b_l \in  \mathbb{R}^{N_{l+1}} $ are biases. The function $\phi$ is called \emph{activation function}, and it is applied component-wise to each component of the vector $y_l$. This function must be nonlinear, and usually is required to be monotonically nondecreasing, continuously differentiable (almost everywhere) and, often, bounded.
Common choices of $\phi$ are sigmoid, hyperbolic tangent, and ReLU.

Given a training dataset $ \mathcal{D}= \{ (X(i), Y(i))  \} $ of input-output sample pairs, 
training of model \eqref{eq:v}-\eqref{eq:FFN} is achieved by tuning its parameters, i.e. all the matrices of weights $W_l$ and vector of biases $b_l$, such that the ``distance'' between the computed output $x_L(i)$ corresponding to the input $ X(i) $, and the desired output $Y(i)$ is as low as possible on the overall training set. 
This translates into minimising
the sum of the distances on each input-output sample, $ \mathcal{L}(\mathcal{D}) = \frac{1}{\text{\#samples}} \sum_i \mathcal{L}(Y(i),x_L(i))$; such distance is commonly called the \emph{loss function}, in this manuscript it will be denoted as $\mathcal{L}$.
Common choices are given by the Mean Squared Error when dealing with regression tasks or the Cross Entropy with classification tasks. The former, for a given input-output sample, reads
\begin{equation*}
    \mathcal{L}(Y(i),x_L(i)) =  \lVert  Y(i) - x_L(i) \rVert ^2,
\end{equation*}
where $\lVert a \rVert$ denotes the Euclidean norm of the vector $ a $.

Note that the computed output $x_L(i)$ is a function of all the parameters of the model and the input $X(i)$, with a nested dependence on the parameters which increases as the depth of the model.
Therefore, considered that the data points $(X(i), Y(i))$ are fixed, we can actively optimise the loss as a function of the parameters of the model, $\mathcal{L} = \mathcal{L}( W_0,b_0, \ldots, W_{L-1},b_{L-1})$.
At a first order approximation, each parameter contributes to the loss in proportion to the derivative of the loss function w.r.t. it.
Thus, in order to minimise the loss, the core idea is to modify each parameter moving towards the opposite direction pointed by the gradients.

The derivative of the mean squared error loss w.r.t. the output is the following vector
$$
    E(i) = - 2(Y(i) - x_L(i))  .
$$
This will be called the \emph{error vector} of the specific input-output sample $(X(i), Y(i))$. However, in the following it will be often denoted shortly as $E$, i.e. dropping the relation with the specific input-output sample.

The derivative of the loss function w.r.t. the weights connecting the $s$-th layer to the $(s+1)$-th layer can be compactly written as the following outer product\footnote{Outer product of vectors $a\in \bR^N$ and $b \in \bR^M$ is a $N\times M$ matrix of 1-rank defined as $ (a \otimes b)_{ij} = a_i b_j $. Equivalently, considering $a, b$, respectively matrices of dimensions $(N,1),(M,1),$ then using the standard matrix multiplication operation, $ a \otimes b = a b^T $. }
\begin{align}
    \label{eq:deriv}
    \dfrac{\partial \mathcal{L}}{ \partial W_{s}} (i)  
    &= 
    \Bigl[ \diag(d_s)  \mathcal{B}_s \bigl( E(i) \bigl)     \Bigl] \otimes \,\, x_s , \qquad s=L-1, \ldots, 0,
\end{align}
where $\diag(d_{s})$ is the diagonal matrix whose diagonal is the following vector
\begin{equation}
    \label{eq:derivative_activation}
    d_s=\phi'(y_s),
\end{equation}
and the linear operator $\mathcal{B}_{s}$ backpropagates the error vector from the output layer up to the $(s+1)$-th layer, as follows
\begin{equation}
    \label{eq:back_operator}
     \mathcal{B}_s\bigl(E(i)\bigl) =     \biggl(\dfrac{\partial x_{L}}{ \partial x_{s+1} } \biggl)^T  E(i) ,
\end{equation}
where the subscript $^T$ denotes the transpose.
The matrix $  \dfrac{\partial x_{L}}{ \partial x_{s+1} }  $ can be expressed as the product of $L-s-1$ matrices,  
\begin{align}
    \label{eq:chain} 
    \dfrac{\partial x_{L}}{ \partial x_{s+1}}  &=  \dfrac{\partial x_{L}}{ \partial x_{L-1}}   
    \dfrac{\partial x_{L-1}}{ \partial x_{L-2}} 
    \ldots
    \dfrac{\partial x_{s+2}}{ \partial x_{s+1}} \\
    \label{eq:deriv_consecutive} 
    \dfrac{\partial x_{l+1}}{ \partial x_{l}} &= \diag(d_l) W_{l} .
\end{align}

Derivatives of the loss function w.r.t. biases $b_s$ can be obtained simply by omitting the outer product with the vector $x_s$ in \eqref{eq:deriv}.
In gradient-based learning methods the weights $W_s$ are updated by subtracting the matrix \eqref{eq:deriv} rescaled by the learning rate $\mu$, that is
\begin{align}
    \label{eq:update}
    W_s(i) &= W_s(i-1) + \Delta W_s (i);\\
    \label{eq:correction}
    \Delta W_s (i) &= - \mu  \dfrac{\partial \mathcal{L}}{ \partial W_{s}}(i) .
\end{align}
In order to accelerate the learning process, often training is accomplished dividing the dataset in batches of $m$ training samples $ \mathcal{D}_j = \bigl\{ \bigl(X(j),Y(j)\bigl), \ldots, \bigl(X(j+m-1), Y(j+m-1)\bigl) \bigl\} $. In that case the correction matrix \eqref{eq:correction} is averaged over a batch as follows
\begin{equation}
    \label{eq:batch_update}
    \Delta W_s (j) = -  \dfrac{\mu}{m} \sum_{i=0}^{m-1} \dfrac{\partial \mathcal{L}}{ \partial W_{s}}(j+i) .
\end{equation}
Whether through mini-batches \eqref{eq:batch_update} or one-by-one samples \eqref{eq:correction}, the update rule \eqref{eq:update} realises an SGD algorithm leading to a journey in the parameter space that will converge to a (hopefully low enough) minimum point of the loss function.

\subsection{The V/E gradient problem.}
\begin{defn}
\label{def:inpout_jacobian}
Let be given an input $x_0=X(i)$, then it is defined through \eqref{eq:v}-\eqref{eq:FFN} the sequence of preactivations $y_1, \ldots, y_{L-1}$ corresponding to $X(i)$, and so the sequence of derivatives $d_1, \ldots, d_{L-1} $ via \eqref{eq:derivative_activation}. It is well-defined via \eqref{eq:deriv_consecutive} the product matrix 
\begin{equation}
    \label{eq:input_output_jacobian}
    \dfrac{\partial x_{L}}{ \partial x_{1} } (i) =  \dfrac{\partial x_{L}}{ \partial x_{L-1}} \dfrac{\partial x_{L-1}}{ \partial x_{L-2}} \ldots \dfrac{\partial x_{2}}{ \partial x_{1}}
    = \prod_{l=L-1}^{1}  \diag(d_l) W_{l} 
    ,
\end{equation}
which is called the \emph{input-output Jacobian} (IOJ) of the NN evaluated on the input $X(i)$.
\end{defn}
\begin{remark}
Note that the linear operator mapping the error $E(i)$ backwardly to the first layer is nothing but the transpose of the IOJ. In fact, eq. \eqref{eq:back_operator} for $s=0$ reads $ \mathcal{B}_0\bigl(E(i)\bigl) = \Bigl( \dfrac{\partial x_{L}}{ \partial x_{1} } (i) \Bigl)^T E(i) = \Bigl( \prod_{l=1}^{L-1} W_{l}^T  \diag(d_l) \Bigl) E(i)  $.
\end{remark}

The IOJ is a product of $L-1$ matrices. Similarly to the product of $L-1$ real numbers, as the depth $L$ increases, the IOJ typically tends to have norm either progressively closer to zero or exploding to infinity, regardless of the input $X(i)$.
This, respectively, vanishes the update rule \eqref{eq:update} making the learning dynamics effectively immobile for those weights far from the output, or make it explode provoking numerical overflows.
In both cases, training very deep NNs becomes unfeasible.

\subsection{The proposed solution} 
The key idea is to realise between all layers a convex combination between the standard MLP transformation and a semi-orthogonal transformation.
In formula, model \eqref{eq:v}-\eqref{eq:FFN} is modified by adding random semi-orthogonal matrices $O_l$ as follows
\begin{equation}
    \label{eq:proposed_model}
    x_{l+1}  = \alpha\phi( W_l x_l + b_l ) \,\, + \,\, (1-\alpha) O_l x_l, \qquad l=0,1, \ldots , L-1,
\end{equation}
where $\alpha$ is a parameter to set accordingly to the depth of the model, and $O_l$ are randomly generated semi-orthogonal matrices with dimensions matching those of $W_l$.
Model \eqref{eq:proposed_model} is called a \emph{random orthogonal additive Feedforward NN (roaFNN)}.
\begin{defn}
A matrix $A \in \bR^{N\times M}$ is called \emph{semi-orthogonal} if either $ A^T A = I_M $ (when $ M<N$), or $ A A^T = I_N $ (when $ M \geq N $), where $I_n$ is the identity matrix of dimension $n$.
\end{defn}
\begin{remark}
The condition $A^T \,\, A = I_M $ implies that $ \| A x \| = \| x \| $ for all $x \in \bR^M$. In fact, $ \| A x \|^2 = ( A x )^T (A x) = x^T A^T A x = x^T x = \|  x \|^2  $.
Analogously, the condition $A \,\, A^T = I_N $ implies that $ \| A^T x \| = \| x \| $ for all $x \in \bR^N$. In either cases, it holds that $\| A \| = 1 $, denoting $\| A \|$ the matrix norm of $ A $ induced by the Euclidean norm of vectors, i.e. the maximum singular value.\\
\end{remark}

\textbf{How to generate random semi-orthogonal matrices.} There are few ways to generate random semi-orthogonal matrices.
One method is to first generate a random matrix $D$ of the desired dimension $N\times M$, hence perform a QR decomposition of $D=QR$, and take the resulting $N\times M$ semi-orthogonal matrix $Q$.
All the simulations in this paper have been run according to this method, generating $D$ with i.i.d. uniformly random entries in $(-1,1)$. However, note that this method does not ensure to explore uniformly the space of orthogonal matrices \cite{mezzadri2006generate}.\\

Since the semi-orthogonal matrices of model \eqref{eq:proposed_model} are kept constant, the derivative of the mapping from layer $l$ to layer $l+1$ w.r.t. weights is simply rescaled by $\alpha $.
Consequently, equation \eqref{eq:deriv} is rescaled as follows\footnote{This rescaling has the consequence to effectively reduce the size of the gradients by a factor of $\alpha$. 
That's the reason why learning rates in the following simulations are higher than usually set in the literature.}
\begin{equation}
\label{eq:back_operator_alpha}
    \dfrac{\partial \mathcal{L}}{ \partial W_{s}} (i)  
    = \alpha
    \Bigl[ \diag(d_s)  \mathcal{B}_s \bigl( E(i) \bigl)     \Bigl] \otimes \,\, x_s , \qquad s=L-1, \ldots, 0,
\end{equation}
where $ \mathcal{B}_s \bigl( E(i) \bigl)  $ is defined as in \eqref{eq:back_operator}, but with the major innovation that the structure of the Jacobian in \eqref{eq:deriv_consecutive} now reads
\begin{equation}
    \label{eq:ortho_consecutive}
    \dfrac{\partial x_{l+1}}{ \partial x_{l}}  = \alpha \,\, \diag(d_l) \,\,  W_l  \,\, + \,\, (1-\alpha)\,\,  O_l .
\end{equation}

Roughly speaking, by a continuity argument when $\alpha$ goes to zero the Jacobian \eqref{eq:ortho_consecutive} tends to the semi-orthogonal matrix $O_l$, and consequently it inherits the isometric property in the limit of $\alpha \rightarrow 0$.
\emph{The insight is that with an appropriate choice of $\alpha$ (reciprocally depending on the depth $L$) we can control the spread of the singular value distribution of the IOJ, thereby promoting approximate dynamical isometry.} 
The challenge is thus to find the sweet spot for $\alpha$ depending on $L$. As stated in the following theorem \ref{thm:main_thm}, whose proof can be found in appendix \ref{sec:bounds}, the choice of $ \alpha = (L-1)^{-1}$, or at least in a neighbourhood of $(L-1)^{-1}$, prevents the IOJ to explode or vanish.\\

\begin{thm}
\label{thm:main_thm}
Let us consider the model \eqref{eq:proposed_model} with an activation function $\phi$ such that its derivative assumes values in $[0,r]$.
Let us denote $\sigma_{\text{max}} := \max_l \lVert W_l \rVert$, i.e. the maximum weights norm among all layers. 
If we parametrise $\alpha = \rho (L-1)^{-1}$, with a positive real valued $\rho$, then the maximum singular value of the IOJ of the model \eqref{eq:proposed_model} admits the following $L$-independent upper bound
\begin{equation}
    \label{eq:upper_bound}
    \biggl\lVert \dfrac{\partial x_{L}}{ \partial x_{1}} \biggl\rVert \,\, \leq \,\, \exp\bigl( \rho(r\sigma_{\text{max}} - 1) \bigl).
\end{equation}\\
Moreover, if $\rho < \dfrac{L-1}{1+r\sigma_{\text{max}}}$, and assuming the sequence of layers' dimensions (i.e. the number of neurons in each layer) from input to output is nonincreasing, 
then it holds the following $L$-independent lower bound
\begin{equation}
    \label{eq:lower_bound}
    \exp\bigl(-\rho(1+r\sigma_{\text{max}})\bigl) \,\, \leq \,\, \biggl\lVert  \dfrac{\partial x_{L}}{ \partial x_{1}} \biggl\rVert.
\end{equation}
\end{thm}

\begin{remark}
Although a proof for the lower bound in the case of increasing sizes of layers could not be found, simulations (not reported) suggest that extremely deep roaFNN models with increasing sizes of layers can be trained effortlessly well as nonincreasing ones. Therefore, the assumption of nonincreasing layers' dimensions is likely to be unnecessary for the lower bound to hold.
\end{remark}
\begin{remark}
Note that, via the parametrisation of $\alpha = \rho(L-1)^{-1} $, the bounds of Theorem \ref{thm:main_thm} hold even in the case of infinite depth $L \rightarrow \infty$.
This implies that even with a prohibitively large number of layers all the parameters of a roaFNN model are updated in a controlled way depending on the three hyperparameters: $\sigma_{max}, r, \rho$.
Moreover, the lower bound \eqref{eq:lower_bound} ensures that even in the case of infinite depth the product of Jacobians will not converge to the null matrix when $L\rightarrow \infty$.
Although Theorem \ref{thm:main_thm} ensures that a standard choice of $\rho=1$ suffices to prevent the IOJ to either vanish or explode, the learning dynamics might be heavily influenced by the particular choice of $\rho$. 
Therefore, for optimal performance the parameter $\rho$ should be optimised for the task at hand.
\end{remark}
%
%
\begin{remark}
The particular case of the semi-identity matrix in place of random semi-orthogonal matrices, corresponds to add a weighted (via $\alpha$) residual connection between each consecutive layer. In this sense, model \eqref{eq:proposed_model} would be a sort of weighted ResNet which however does not make use of skip connections. In section \ref{sec:ortho_role}, comparative experiments between the proposed model with the identity and random orthogonal matrices can be found, w.r.t. recurrent models. 
Similarly, roaFNN is related to Highway Networks \cite{srivastava2015highway}, but it circumvents the need of trainable gates at a structural level. Moreover, neither ResNets nor Highway Networks are provided with formal proofs of solving the V/E gradient issue and, even more so, for the case of arbitrarily large depths. On the other hand, the roaFNN model is rooted on solid mathematical bounds regarding the maximum singular value of the IOJ that still holds in the case of infinite depths.\\
\end{remark}

In summary, via backpropagation of the errors the update rule \eqref{eq:update}-\eqref{eq:correction} modifies the parameters of the model \eqref{eq:proposed_model}, $W_l$ and $b_l$, while the matrices $O_l$ automatically keep the gradient norms at a controlled magnitude.
\emph{The role of those fixed matrices is essentially to ``shuffle'' the information encoded in the neuronal activations at each crossed layer (forwardly or backwardly) encouraging dynamical isometry.}
This simple trick results to be effective for a FNN to achieve approximate dynamical isometry, i.e. a configuration where the distribution of the singular values of the IOJ remains  contained  ``around"  1  without  spreading  too  far  from  it. 
To the best of the author's knowledge, a general method to ensure approximate dynamical isometry in nonlinear neural networks was missing so far.
Remarkably, this is accomplished without imposing any constraints on the parameters of the model or relying on specific initialisation schemes.
Noticeably, a roaFNN is a clear generalisation of a standard MLP, which we recover setting $\alpha=1$, or equivalently setting $\rho=L-1$. Evidently, the bounds of Theorem \ref{thm:main_thm} exponentially diverge to infinity or collapse to zero, in the case of $\rho=L-1$, for increasingly larger depths.\\

\textbf{Computational complexity.}
Due to the presence of the additional semi-orthogonal matrices, both the computational time and memory are doubled in a roaFNN compared to a vanilla FNN.
In any case, considering the outstanding improvement of the performance, especially for RNNs (see section \ref{sec:simulations}), this trade-off is particularly favourable. Furthermore, in Appendix \ref{sec:reducing_memory_time}, two strategies to bring memory and time consumption of model \eqref{eq:proposed_model} to the same levels of the vanilla FNN are proposed. However, those strategies have not been properly tested in this work, therefore further investigations are needed.

\subsection{Double-moon regression task}
\label{sec:double_moon}

\begin{figure}[ht!]
    \centering
    \includegraphics[keepaspectratio=true,scale=0.15]{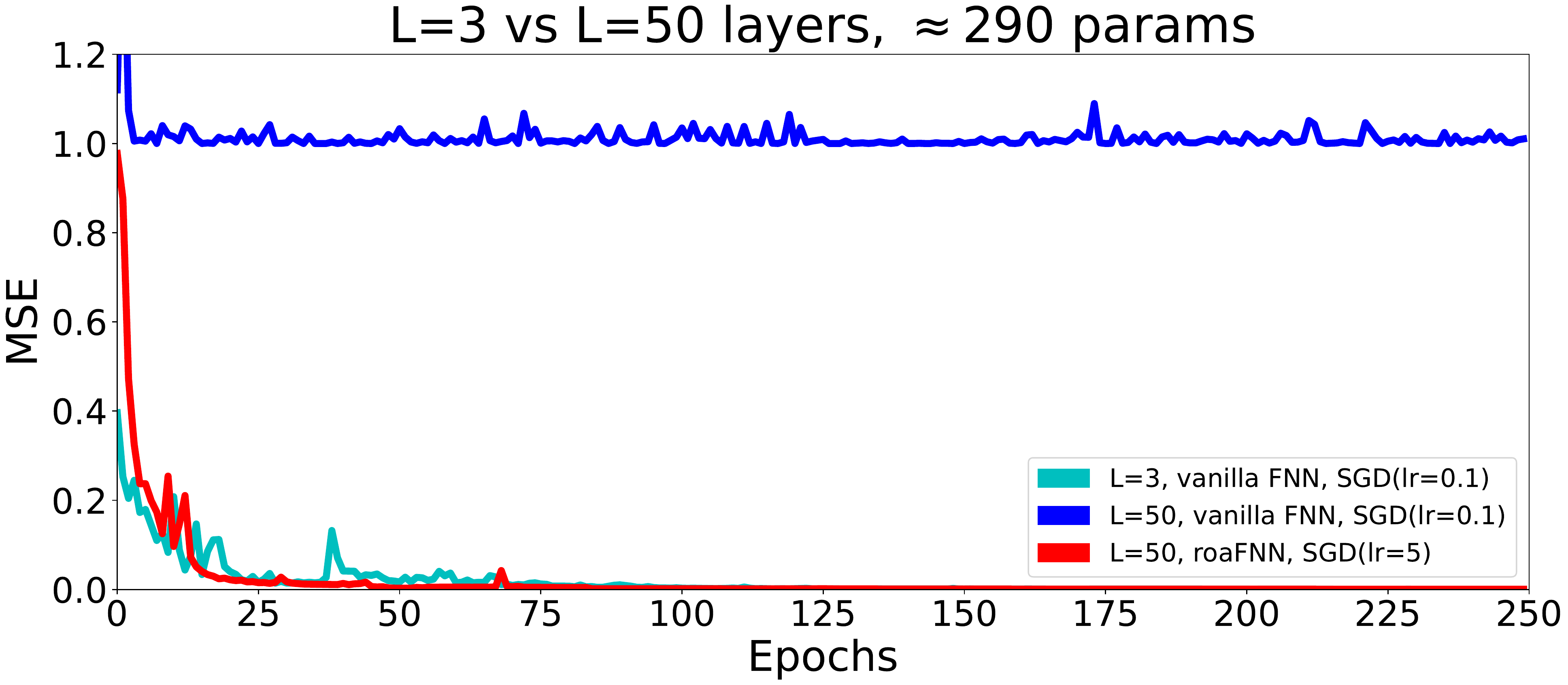}~\hspace{0.3cm}~\includegraphics[keepaspectratio=true,scale=0.15]{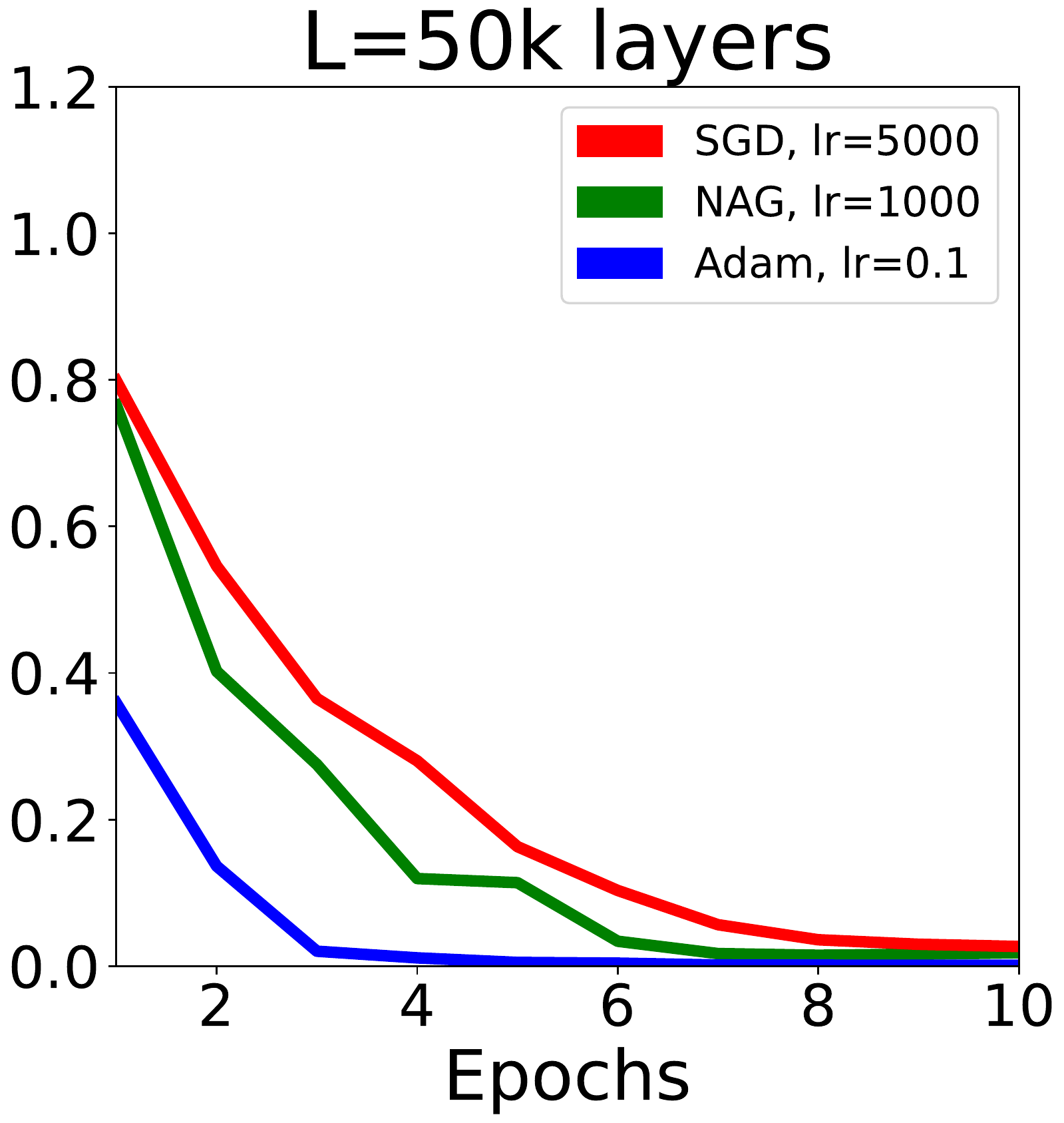}\\
    \vspace{0.15cm}
    \includegraphics[keepaspectratio=true,scale=0.21]{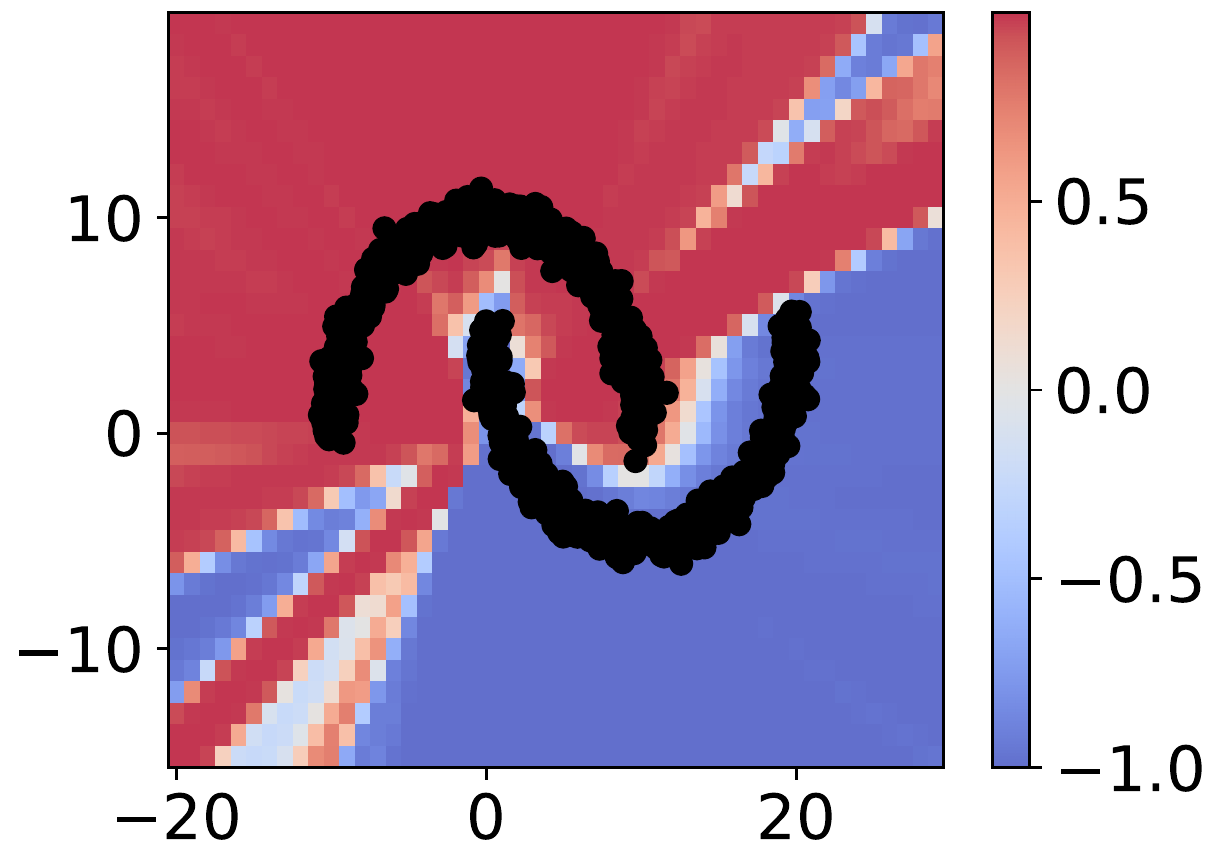}~\hspace{0.4cm}~\includegraphics[keepaspectratio=true,scale=0.21]{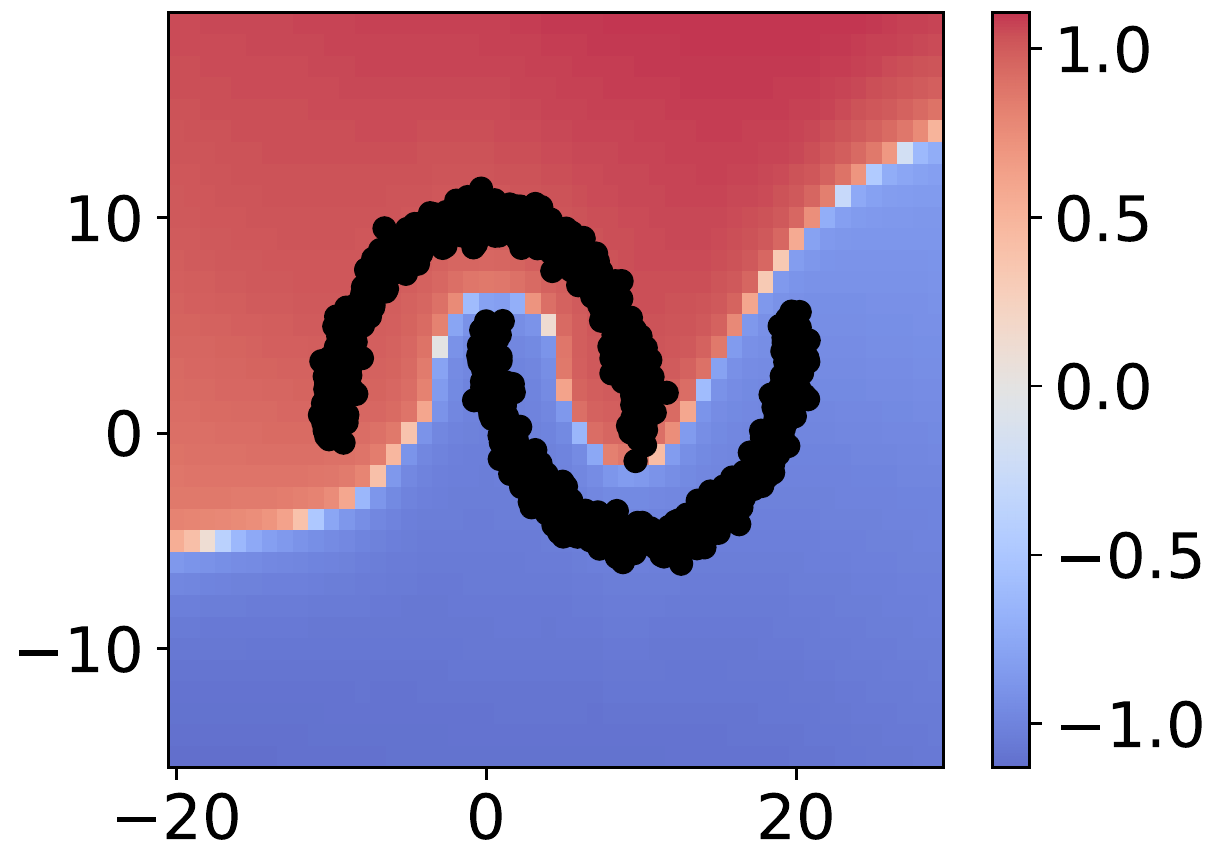}~\hspace{0.4cm}~\includegraphics[keepaspectratio=true,scale=0.21]{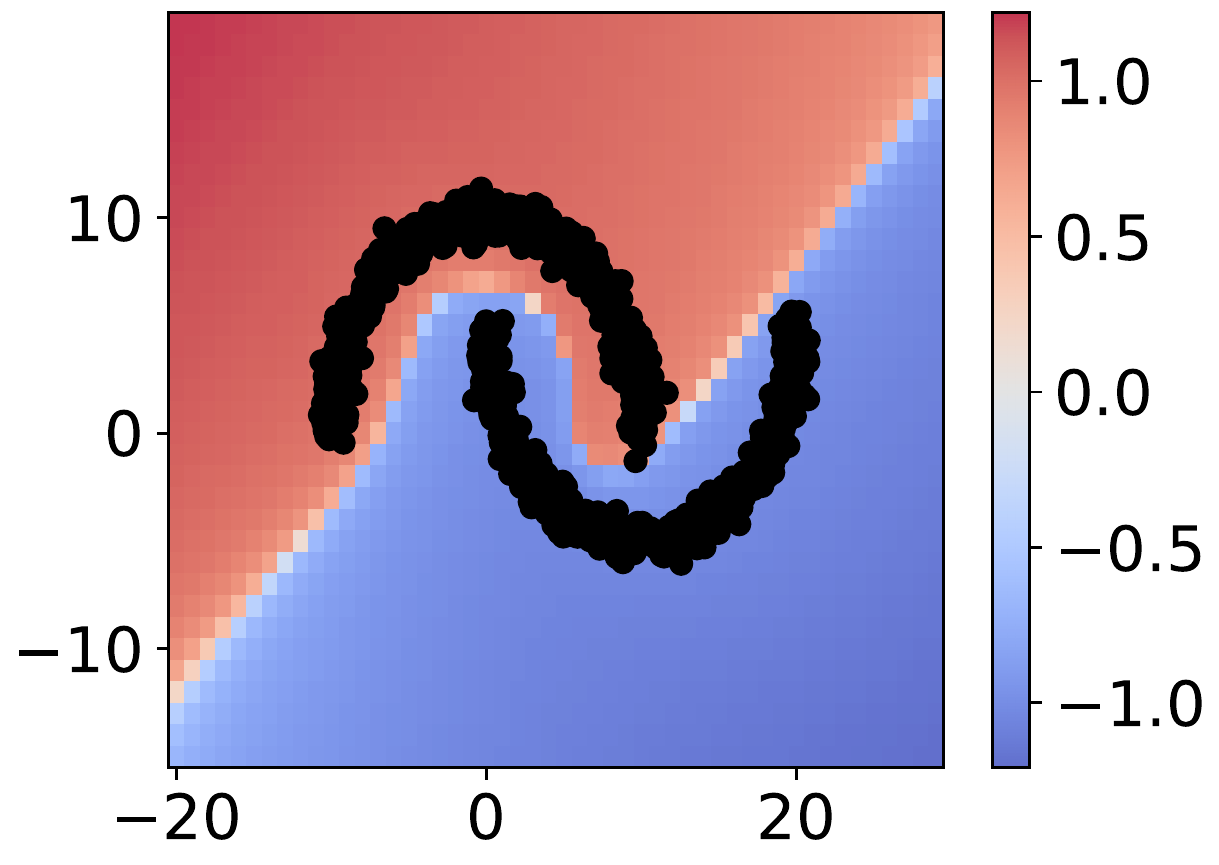}
\caption{\textbf{Top left.} In cyan the MSE of a shallow vanilla FNN with one hidden layer of dimension 72 (trainable parameters 289). In blue, and red, the MSE of, respectively, a deep vanilla FNN, and a deep roaFNN, with 48 hidden layers of dimension 2 (trainable parameters 291).
\textbf{Top right.} A roaFNN of 50k layers has been trained with 3 different optimisers. SGD is stochastic gradient descent (no momentum). NAG is Nesterov accelerated gradient with $0.99$ momentum. Adam is with standard parameters suggested in \cite{kingma2014adam}.
\textbf{Bottom row.} The output of the models, evaluated in the region $(x_1, x_2)\in [-20,30]\times [-15,20]$. 
From left to right: the shallow $L=3$ vanilla FNN trained with SGD for 250 epochs, the deep $L=50$ roaFNN trained with SGD for 250 epochs, and the extremely deep $L=50k$ roaFNN trained with Adam for 10 epochs. The deep $L=50$ vanilla FNN is not reported since it did not solve the task. \textbf{All.} All models have input and hidden layers of dimension 2, and output layer of dimension 1. All neurons have $\tanh$ activation function. All weights and biases have been initialised according to a Normal distribution with zero mean and unitary standard deviation. Batchsize is 100. All roaFNN models have $\alpha = \rho(L-1)^{-1}$ with $\rho=5$. Learning rates (denoted as lr in the in the figure) of roaFNNs are larger than usual, but they are relatively small considering that they are effectively rescaled by a factor of $\alpha$ in the backpropagation algorithm, see eq. \eqref{eq:back_operator_alpha}. Adam is an exception since it autonomously adapts the learning rate.
}
\label{fig:double_moon1}
\end{figure}

In this section, as a proof of concept, it is shown that we can propagate signals back and forth through an extremely deep FNN.
The task is the double-moon regression task \cite{haykin2010neural}. A number of 1000 sample points $(x_1,x_2)$ in the real plane are generated in a double-moon fashion: black dots in the second row plots of Figure \ref{fig:double_moon1}. Points of one moon are labelled with target value $-1$ and the others as $1$. 
Based on the double-moon training dataset, an FNN is asked to generalise a mapping from $\bR^2$ to $\bR$. 
It is a pretty trivial regression task for a shallow vanilla FNN with one hidden layer of 72 neurons.
Nevertheless, for a vanilla FNN with the same number of trainable parameters spread over 48 hidden layers ($L=50$), it turns out impossible. On the contrary, a 50-layers roaFNN converges fast and smoothly to a solution in few epochs. The same rate of convergence is achieved with a 50000-layers roaFNN, with no gradient clipping or any other ``trick'' (e.g. batch normalisation, skipping connections, etc).
This corroborates the resilience of the proposed model to V/E gradient issues. For details of the simulations see caption of Figure \ref{fig:double_moon1}.

Unlike the vanilla FNN, the roaFNN model presents a smooth decision boundary, even for the 50k layers model despite the abnormally large number of parameters to tune ($\approx 300k$).
The smoothness of the learning curve and the decision boundary for roaFNN are likely due to the presence of the orthogonal matrices in \eqref{eq:proposed_model} that may have the effect to regularise the landscape of the loss function and stabilise the learning dynamics; note that no explicit regularisation has been introduced in any model.
Both strategies (I) and (II) proposed in appendix \ref{sec:reducing_memory_time} have been tested on this task, and they both worked fine (data not reported).

\section{Learning long-term dependencies with RNNs}
\label{sec:}

The idea presented for FNNs in section \ref{sec:feedforward} can be implemented in any kind of neural network model. In this section, I will implement the idea on \emph{vanilla RNNs}.
Generally, an RNN processes an input sequence $\{ u_k \}$, developing an internal state sequence $ \{ x_k \}$, and generating an output sequence $\{ z_k \}$, by means of the following state-update equations 
\begin{align}
\label{eq:hiddenRNN}
x_{k+1} & = \phi( W_h x_k + b_h + W_i u_{k+1} ), \\
\label{eq:outputRNN}
z_{k+1} & = \psi( W_o x_{k+1} + b_o ) .
\end{align}
Typical choices of $\psi$ are the identity function, or a softmax function.

Starting with an initial condition $x_0$ for the internal state, and considering a finite window of evolution of $L$ time steps, we can interpret \eqref{eq:hiddenRNN}-\eqref{eq:outputRNN} as a special FNN driven by the finite input sequence $ (u_1, u_2, \ldots , u_L) $. 
Therefore, we can fictitiously unfold the RNN in an FNN with (internal) input layer $x_0$ passing through $L$ hidden layers that share the same weights and biases, $ W_h $, $b_h$ and driven by an (external) input sequence $ (u_1, u_2, \ldots , u_L) $ via further shared parameters $W_i$.
Hence, we can use the backpropagation algorithm to update the parameters of model \eqref{eq:hiddenRNN}-\eqref{eq:outputRNN}. This is called the \emph{backpropagation through-time} algorithm.
As a consequence, such an RNN must be able to learn dependencies in the input sequence for up to $L$ time steps in the past in order to fuction correctly.
However, when it comes to learn long-term dependencies, a vanilla RNN model of equations \eqref{eq:hiddenRNN}-\eqref{eq:outputRNN} suffers terribly from the V/E gradient issue.
For example, this can be easily noticed from the fact that the IOJ for a vanilla RNN reads $ \prod_{l=L-1}^1 \diag(d_l) W_h $, whose norm can be upper-bounded with $ \lVert W_h \rVert^{L-1} $, whenever the activation function assumes derivative values in $[0,1]$. Hence, for large values of $L$, if $ \lVert W_h \rVert < 1 $, we observe vanishing gradients.

A simple way to enhance the computational power of a vanilla RNN is suggested by the arguments discussed in section \ref{sec:feedforward}. The proposed RNN model is the following:
\begin{align}
\label{eq:ortho_hiddenRNN}
x_{k+1} & = \alpha \phi( W_h x_k + b_h + W_i u_{k+1} ) + (1-\alpha) O x_k , \\
\label{eq:ortho_outputRNN}
z_{k+1} & = \psi( W_o x_{k+1} + b_o ) ,
\end{align}
where $O$ is a random orthogonal matrix.
In the remainder of this paper, model \eqref{eq:ortho_hiddenRNN}-\eqref{eq:ortho_outputRNN} will be referred to as roaRNN, which stands for Random Orthogonal Additive RNN.
In summary, we simply add a random orthogonal matrix shared between ``all the hidden layers" of an RNN.
Note that in the limit of $\alpha=1$ we recover the vanilla RNN model.

\begin{remark}
The orthogonal matrix $O$ characterises the internal redistribution of the information into the network.
The matrix $(1-\alpha)O$ has eigenvalues located in the unit disk, close to the unit circle for smaller values of $\alpha$.
Such close-to-orthogonal matrix represents the slightly dissipative internal dynamics of the recurrent network.
Importantly, an $0<\alpha<1$ ensures that the dynamical system defined by \eqref{eq:ortho_hiddenRNN} evolves inside the hypercube $[-R, R]^{N_h}$, as long as the activation function $\phi$ is bounded in $(-R,R)$, non-decreasing, and upper semi-continuous, see \cite[Proposition B.1]{ceni2020echo} for a proof of this fact.
\end{remark}
 

\subsection{RoaRNN gradients}
\label{sec:RNN_bounds}

Parameters of model \eqref{eq:ortho_hiddenRNN}-\eqref{eq:ortho_outputRNN} are $W_h , b_h, W_i, W_o , b_o$.
We want to adapt parameters of the model proportionally to the derivative of the loss function w.r.t. them.

Gradients of ``hidden'' parameters, i.e. $ W_h , b_h, W_i $, need special attention since weights are shared across temporal layers.
For instance, the gradient w.r.t. the hidden-to-hidden weights takes the form
\begin{equation}
\label{eq:hidden_gradients}
    \dfrac{\partial \mathcal{L}}{ \partial W_{h}} =  \sum_{s=0}^{L-2} \dfrac{\partial \mathcal{L} }{ \partial W_{h,s}} ,
\end{equation}
where the same hidden-to-hidden matrix $W_h$ has been fictitiously labelled as $W_{h,s}$, as if they were different matrices at each temporal layer.
For each $s=0, \ldots, L-2$, it holds
\begin{align*}
    \dfrac{\partial \mathcal{L} }{ \partial W_{h,s}} &= \alpha \Bigl[  \diag\bigl(\phi'(y_s)\bigl) \mathcal{R}_s(E) \Bigl] \otimes x_s \\
    \mathcal{R}_s(E) &= \Bigl(\dfrac{\partial x_L }{ \partial x_{s+1}} \Bigl)^T  \Bigl( \diag\bigl(\psi'(y_L)\bigl) W_o \Bigl)^T E\\
    \dfrac{\partial x_L }{ \partial x_{s+1}} &=  \dfrac{\partial x_{L}}{ \partial x_{L-1}}
    \dfrac{\partial x_{L-1}}{ \partial x_{L-2}}
    \ldots
    \dfrac{\partial x_{s+2}}{ \partial x_{s+1}}, 
\end{align*}
where $ E $ is the error vector, $y_l = W_h x_l + b_h + W_i u_{l+1}  $, and 
\begin{equation}
    \label{eq:rnn_jacobian}
    \dfrac{\partial x_{l+1}}{ \partial x_{l}}
    = \alpha  \diag(\phi'(y_l))  W_{h}   +  (1-\alpha)  O   
\end{equation}

The analysis of roaRNN gradients is much easier than roaFNNs since we deal with a square IOJ matrix.
In fact, we can bound completely the distribution of singular values of the IOJ, as stated in the following theorem. \\

\begin{thm}
\label{thm:distribution_singvals}
Let us consider a roaRNN with an activation function $\phi$ such that its derivative assumes values in $[0,r]$. Let be given any initial state $x_0$, driven by the input sequence $( u_1, \ldots, u_L)$. Therefore, via \eqref{eq:ortho_hiddenRNN} it is well-defined the input driven trajectory $ \{ (x_0, u_1), \ldots, (x_{L-1}, u_L) \}$.
Denote $\sigma = \lVert W_h \rVert$.
Assume $\alpha=\rho (L-1)^{-1}$, such that $\rho<\dfrac{L-1}{1+r \sigma}$, then the whole set of singular values of the IOJ matrix $ \dfrac{\partial x_L}{\partial x_1}  $ is contained in the $L$-independent interval $$
\biggl[\exp\bigl(-\rho(1+r \sigma)\bigl) , \,\,\, \exp\bigl(\rho(r \sigma-1) \bigl) \biggl].
$$
\end{thm}

The upper bound on the maximum singular value is a straightforward application of \eqref{eq:upper_bound} of Theorem \ref{thm:main_thm} for the case of roaRNN.
The proof of the lower bound on the \emph{minimum} singular value can be found in Appendix \ref{sec:minimum_sing_val}.\\

Putting all together in \eqref{eq:hidden_gradients} we get the expression
\begin{equation}
\label{eq:hidden_grad2}
    \dfrac{\partial \mathcal{L}}{ \partial W_{h}}  =   \alpha\sum_{s=0}^{L-2}   \Bigl[  \diag\bigl(\phi'(y_s)\bigl) \mathcal{R}_s(E) \Bigl] \otimes x_s
    .
\end{equation}
The gradients w.r.t. the other hidden parameters $b_h$, and $W_i$, have the same form of \eqref{eq:hidden_grad2}, but, respectively, omitting the outer products with the vectors $x_s$, and replacing the outer products with the vectors $x_s$ with the vectors $u_{s+1}$.
\begin{remark}
The matrix \eqref{eq:hidden_grad2} is a sum of $L-1$ terms; each one depending on a different depth of the internal representation $\{ x_k \}$ of the listened input sequence, and the input sequence itself.
Because of the parametrisation $\alpha=\rho (L-1)^{-1}$, the sum \eqref{eq:hidden_grad2} is effectively a sort of average of all the contributions from the past $L$ time steps.
\end{remark}

\begin{remark}
Note that Theorem \ref{thm:distribution_singvals} still holds with a roaRNN where the orthogonal matrix is the identity matrix.
Interestingly, such a model has been already proposed in the deep learning literature with the name of FastRNN \cite{kusupati2019fastgrnn}, and in the reservoir computing literature with the name of leaky-integrator ESN \cite{jaeger2007optimization}.
Remarkably, theoretical results on convergence and generalisation error bounds exist for FastRNN \cite[Theorems 3.1-3.2]{kusupati2019fastgrnn} whose proofs automatically hold for the roaRNN architecture.
However, experiments suggest that the identity does not seem to provide performance as good as a random orthogonal matrix, see section \ref{sec:ortho_role}.
The conjecture is that the characteristic annular distribution of the eigenvalues of the Jacobians \eqref{eq:rnn_jacobian} may play a key role. In fact, the eigenvalues of $  \alpha \diag(\phi'(y_l))  W_{h}   +  (1-\alpha)  O $ can be tuned arbitrarily close to the complex unitary circle as $\alpha$ approaches zero, regardless of the matrix $W_h$. On the other hand, replacing $O $ with the identity matrix makes the eigenvalues converge in the complex plane to the point $(1,0)$ as $\alpha \rightarrow 0$. This might negatively affect the expressive power of the model. 
Further comparative experiments of identity vs random orthogonal can be found in these preprints \cite{oaesn2022neuralnet2022,ceni2022esann}, from the point of view of reservoir computing.

A continuous-time version of roaRNN, and its link with the LipschitzRNN model \cite{erichson2020lipschitz}, is discussed in Appendix \ref{sec:CTrandorthoRNN}.
\end{remark}

\section{Experiments on RNNs} 
\label{sec:simulations}

In this section, the roaRNN model is benchmarked on three tasks specifically devised to be challenging for recurrent models to learn long-term dependencies \cite{arjovsky2016unitary}:
\begin{itemize}
    \item[(A)] copying memory (MemCop)
    \item[(B)] adding problem (AddProb)
    \item[(C)] permuted sequential MNIST (psMNIST).
\end{itemize}

The roaRNN model is compared against vanilla RNN\footnote{The vanilla RNN model used here is a slight variation of equation \eqref{eq:hiddenRNN} where there are 2 bias vectors to learn. This is the default PyTorch implementation of an RNN at present.} with ReLU activation function, and LSTM-RNN, which is by far the most popular RNN model used to learn long-term dependencies.\\

\textbf{RoaRNN setting.} For all the considered tasks, all parameters $W_h, b_h, W_i, W_o, b_o$, are initialised according to a normal distribution with zero mean and unitary standard deviation. The activation function $\phi$ is ReLU.
The readout function $\psi$ in \eqref{eq:ortho_outputRNN} is chosen as the identity, that is a linear output layer. The learning rate, after random trials in the range $[ 10^{-4}, 10 ]$, has been set to $0.5$, and kept constant throughout the learning for the MemCop and AddProp tasks, while for the psMNIST it has been set to $0.1$ in the first 10 epochs, and then to $0.01$ for the next 10 epochs. The hyperparameter $\rho$ is optimised on each task on the grid of values $\{ \frac{1}{300} , \frac{1}{200} , \frac{1}{100} , \frac{1}{50} , \frac{1}{10}, \frac{1}{2},  1, 2, 3, 5, 10, 50 \}.$

\textbf{Vanilla RNN setting.} The hidden-to-hidden matrix is initialised as an orthogonal matrix. All the other parameters are initialised according to a uniform distribution in $(-\sqrt k, \sqrt k)$ where $k = N_h^{-1}$ is the reciprocal of the hidden size. The activation is ReLU. The readout function $\psi$ in \eqref{eq:ortho_outputRNN} is chosen as the identity, that is a linear output layer. The learning rate for the MemCop and AddProb tasks, after random trials in the range $[ 10^{-6}, 1 ]$, has been set to $10^{-4}$, and kept constant throughout the learning. 

\textbf{LSTM setting.} All the 4 hidden-to-hidden matrices are initialised as orthogonal matrices. All the other parameters are initialised according to a uniform distribution in $(-\sqrt k, \sqrt k)$ where $k = N_h^{-1}$ is the reciprocal of the hidden size. As usual, the activation is $\tanh$, while gates are implemented with sigmoids. The readout function $\psi$ in \eqref{eq:ortho_outputRNN} is chosen as the identity, that is a linear output layer.
The learning rate for the MemCop and AddProb tasks, after random trials in the range $[ 10^{-6}, 1 ]$, has been set to $5\cdot 10^{-3}$, and kept constant throughout the learning. 

\textbf{Optimiser.} All models are trained with the Adam optimiser, according to the standard setting proposed in \cite{kingma2014adam}. No regularisation, no gradient clipping, no batch normalisation, were used in any model.

\textbf{Hardware.} CPU is an 11th Gen Intel Core i7-11370H. GPU is a 4GB NVIDIA GeForce GTX 1650.

\textbf{Software.} Experiments have been run with pytorch 1.10.1, and cudatoolkit 10.2.89.

\textbf{Trials.} In order to wash out unlucky initialisations from the comparison, 5 randomly generated seeds between $0$ and $2^{30}$ have been used to run each model. The plots regarding the MemCop and AddProb tasks refer to the best results obtained by each model out of the 5 trials.

\textbf{Results.} Remarkably, roaRNN reaches state-of-the-art performance on all tasks, despite the minimal change from a vanilla RNN model\footnote{Different combinations of initialisations, e.g. all parameters uniformly random in $(-1,1)$, optimisers, e.g. SGD, and activation functions, e.g. $\tanh$, work good as well. However, hyperparameters as $\rho$ and the learning rate should be changed accordingly for optimal performance.}.

\subsection{Copying memory task}
\label{sec:copy_mem}
\begin{figure*}[ht!]
    \centering
    \includegraphics[keepaspectratio=true,scale=0.24]{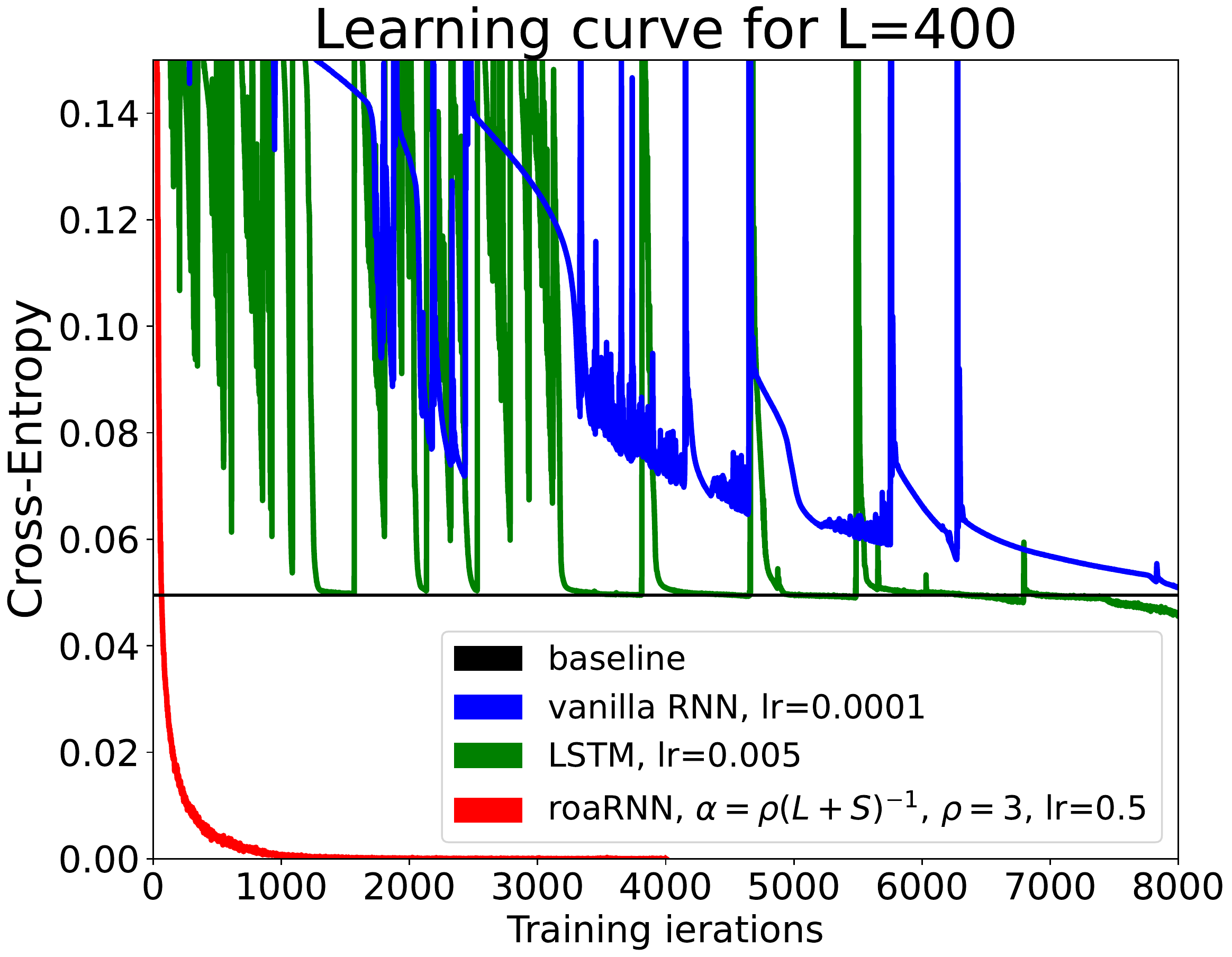}~\includegraphics[keepaspectratio=true,scale=0.24]{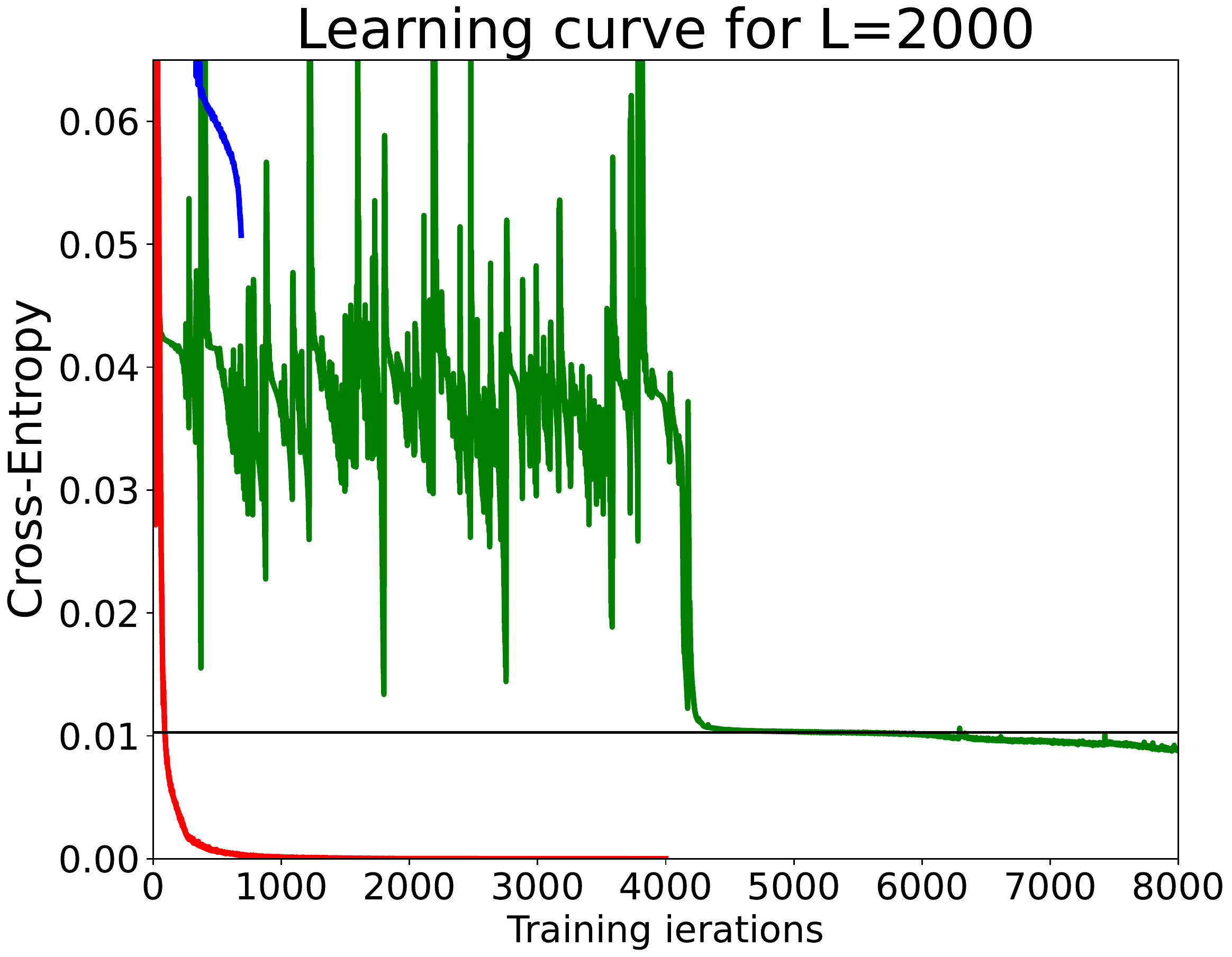}~\includegraphics[keepaspectratio=true,scale=0.24]{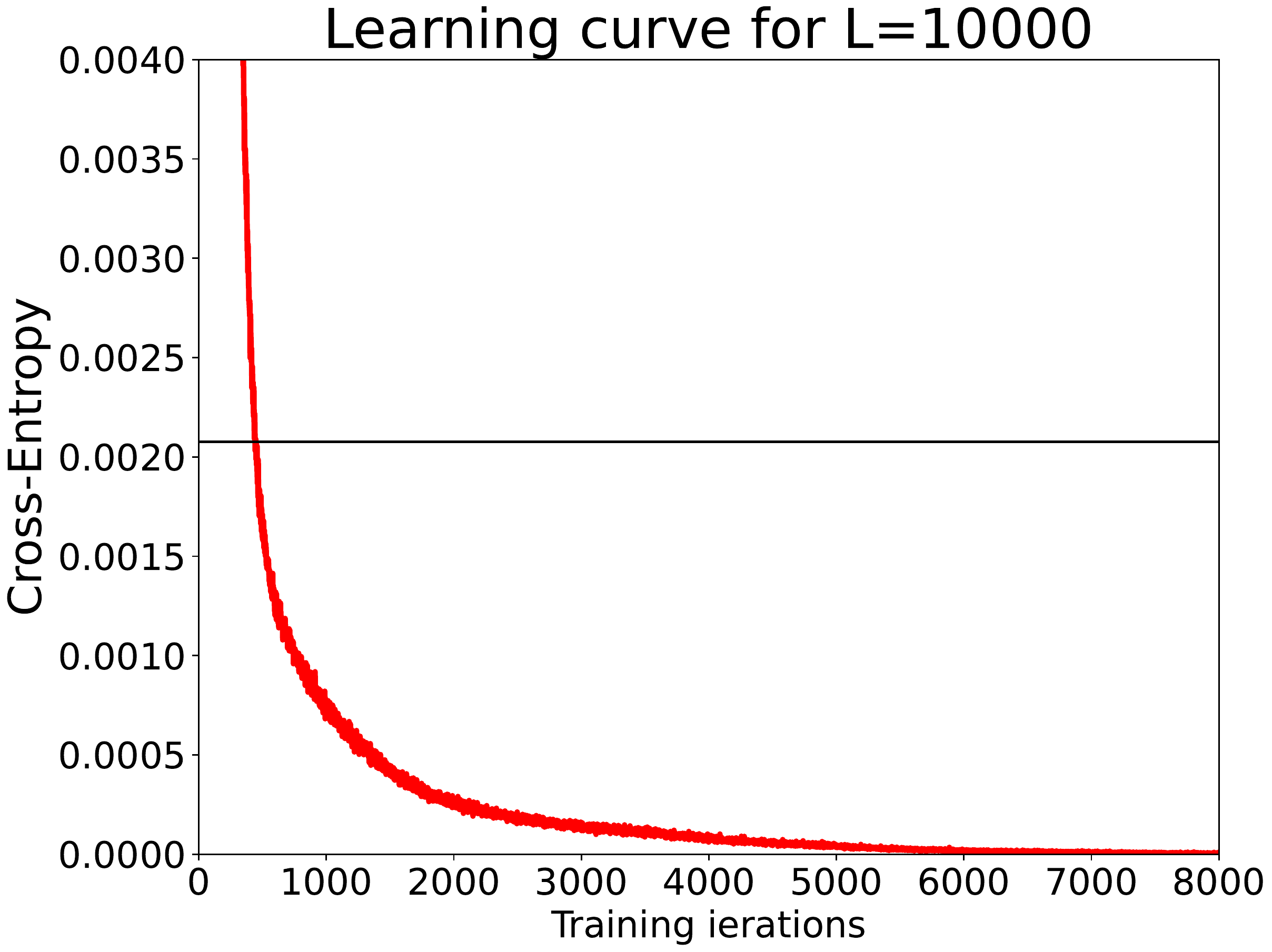}
\caption{Copying memory task. There are $S=10$ symbols to recall after a time lag of $L$ time steps. Cross-entropy loss of vanilla RNN (blue), LSTM (green), roaRNN (red), for three cases of time lags $L=400, 2000, 10000$. In all three cases, roaRNN is set with $\alpha = \rho(L+S)^{-1}$, with a $\rho=3$. In the legend, $lr$ denotes the learning rate.}
\label{fig:copy_mem}
\end{figure*}

\textbf{Task.} Let be given an alphabet of 8 symbols, that we can codify as $ \mathcal{A} = \{ 1, 2, 3, 4, 5, 6, 7, 8 \}$. The input is a sequence composed by 3 concatenated blocks of lengths $(S,L,S)$. The first $S$ time steps are symbols sampled i.i.d. uniformly at random from the alphabet $\mathcal{A}$. Then, there are $L$ time steps of blank characters that we can codify as $0$. Finally, there is a start character codified as $9$ followed by $S-1$ blank characters. The RNN must generate an output sequence of the same length of the input, where the first $S+L$ outputs are blank characters, and the last block of $S$ time steps must match the first block of ordered symbols of the input sequence. The length of the block of symbols to recall is $S=10$, as usually set in the literature \cite{lezcano2019cheap, arjovsky2016unitary, wisdom2016full}. Below an example of input-output pairs with $L=12$ and $S=4$, where the blank character $0$ has been replaced with $-$, and the start character $9$ with $:$, for an easier visualisation.
\begin{align*}
    \textbf{Input \quad} & \,\,5\,\,\,5\,\,\,6\,\,\,1\,------------\,:\,--- \\
    \textbf{Output \quad } &  ----------------\,\,5\,\,\,5\,\,\,6\,\,1 
\end{align*}

In the simulations run, time lags of $L=400, 2000, 10000,$ time steps have been considered.
Batches of size 128 are randomly generated on the fly as training proceeds.
The loss function is the cross-entropy. The baseline to beat for this task is a cross-entropy value of $S \log(8)/(L + 2S)$, corresponding to output $L+S$ blank characters followed by $S$ symbols at random.\\

\textbf{Optimal $\rho$ setting.} Note that different values of $L$ gives different optimal values of $\rho$. However, the setting of $\alpha = \rho(L+S)^{-1}$, with a $\rho=3$ behaves quite well for all the three considered cases of $L=400, 2000, 10000$.   \\

ExpRNN \cite{lezcano2019cheap} and scoRNN \cite{helfrich2018orthogonal} achieve the best performance in the literature for the MemCop task. They use 190 units in their simulations, so the same number has been used here for all the three models, vanillaRNN, LSTM, roaRNN, for sake of comparison. Note however that an LSTM of 190 hidden units has a considerably larger number of trainable parameters than a vanilla RNN, and roaRNN of the same size. Nonetheless, as clear from Figure \ref{fig:copy_mem} roaRNN abundantly outperforms LSTM. As already reported in the literature, LSTMs truly struggle to solve the MemCop task even with relatively  short time lags of $L=400$, albeit obtaining slightly better performance than vanilla RNNs. On the contrary roaRNNs are characterised by a smooth learning curve, and a fast convergence regardless of the length of the time lag $L$. Even in the extreme case of $L=10000$ the roaRNN model beats the baseline within the first 500 iterations.
In the cases of $L=400$ and $L=2000$, the roaRNN training has been stopped at the 4000-th iteration, since it reached $100\%$ accuracy on the vast majority of the training steps after the 2000-th. Running for longer sometimes gives rise to instabilities.
For the case of $L=2000$, all the 5 trials of vanilla RNNs gave NaN values, testifying a V/E gradient issue. The ones plotted in Figure \ref{fig:copy_mem} are those that got NaN values the latest. 
All the simulations for the MemCop tasks have been run with a batch size of 128, as in \cite{lezcano2019cheap}.

\subsection{Adding problem task}
\begin{figure*}[ht!]
    \centering
    \includegraphics[keepaspectratio=true,scale=0.24]{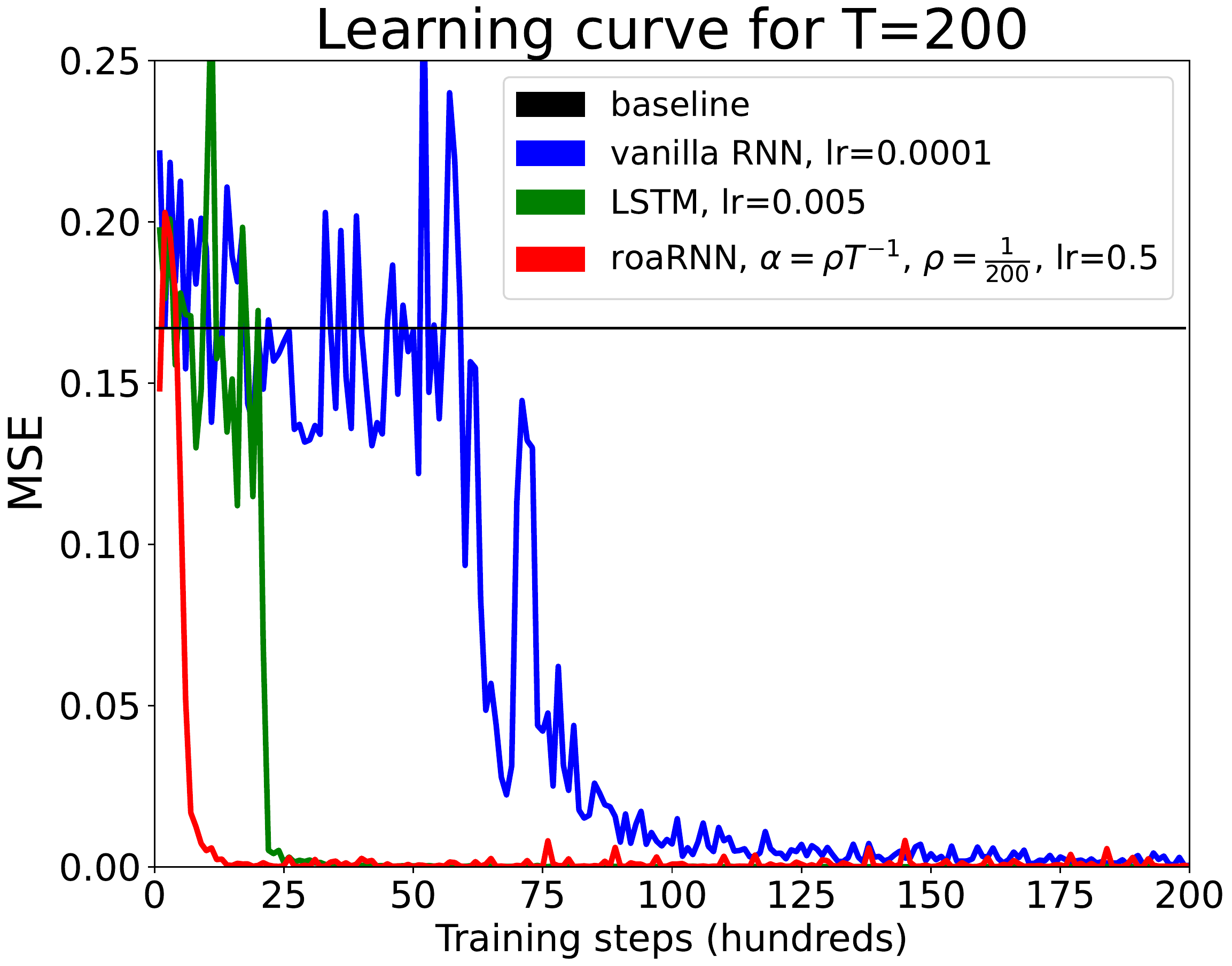}~
    \includegraphics[keepaspectratio=true,scale=0.24]{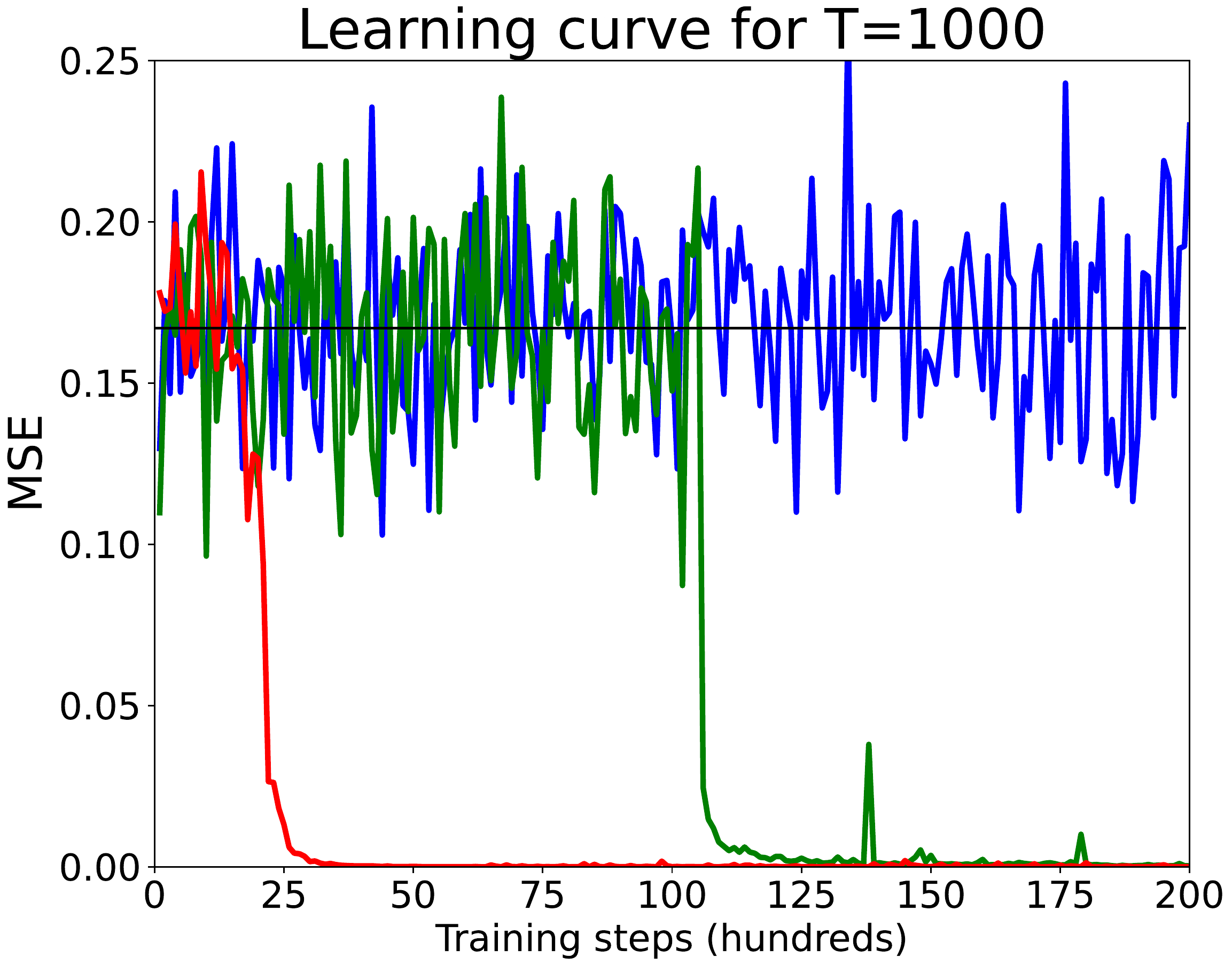}~
    \includegraphics[keepaspectratio=true,scale=0.24]{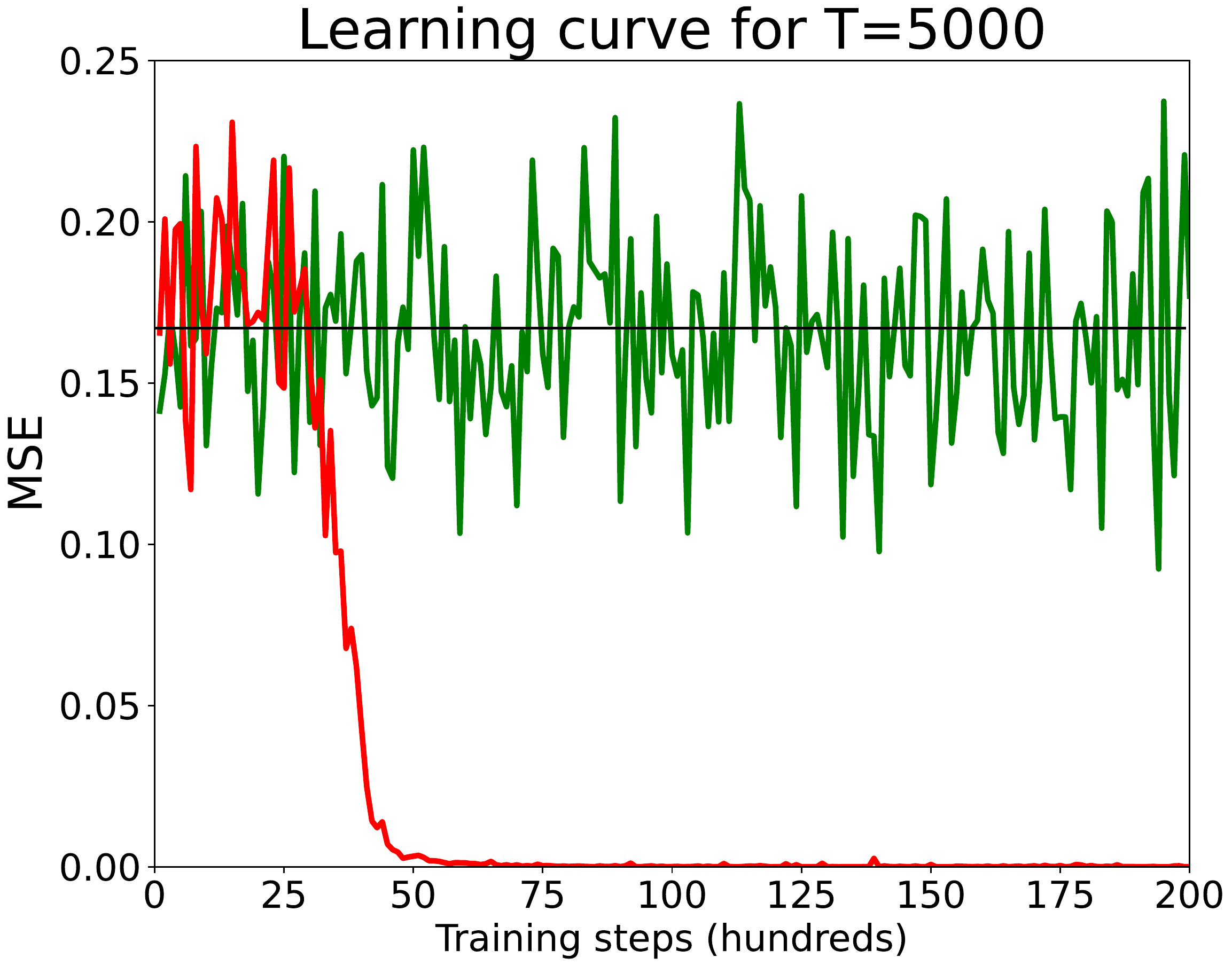}
\caption{Adding problem task. Mean squared error of vanilla RNN (blue), LSTM (green), roaRNN (red), for the three cases of sequence lengths $T=200, 1000, 5000$. In all three cases, roaRNN is set with $\alpha = \rho T^{-1}$, with a $\rho=\frac{1}{200}$. In the legend, $lr$ denotes the learning rate.}
\label{fig:addingproblem}
\end{figure*}

\textbf{Task.} The RNN is driven with 2 input sequences, $u_k = (a_k,b_k)$, of the same length $T$. The first sequence $a_k$ assumes values according to a uniform distribution of real values in $[0,1]$. The second sequence $b_k$ assumes all zeros apart from two time steps, for $k=i_1$ and $k=i_2$, which are marked with the value $1$. The markers $i_1, i_2$, are (uniformly) randomly selected in the first half of the sequence, and the second half of the sequence, respectively, i.e. at positions $1\leq i_1 \leq \lfloor\frac{T}{2}\rfloor < i_2 \leq T $. The RNN must ``recognize'' the two real numbers corresponding to the positions marked with $1$, ignoring the other $T-2$ time steps of uniformly distributed noise, and produce in output the sum of them, i.e. $a_{i_1} + a_{i_2}$.
Below an example of input-output pairs with $T=8$ where the decimal parts are truncated to $10^{-2}$ for an easier visualisation.
\begin{align*}
    \textbf{Input\,\,} & 
        \begin{bmatrix}
        0.71 & 0.14 & 0.64 & 0.77 &
       0.54 & 0.26 & 0.88 & 0.15  \\
        0 & 1 & 0 & 0 &
        0 & 1 & 0 & 0 
        \end{bmatrix}\\
    \textbf{Output\,\,} & \,\,0.14+0.26
\end{align*}

In the simulations run, lengths of $T=200, 1000, 5000$, time steps have been considered.
The loss function for this task is the mean-squared-error.
The baseline MSE value to beat is $0.167$, i.e. the variance of the sum of two independent uniform distributions in $[0,1]$, corresponding to predict the output value $1$ regardless of the input sequence.
Batches of size 50 are randomly generated as training proceeds.
Each hundred of training steps the model is evaluated on a randomly generated batch on the fly and plotted in Figure \ref{fig:addingproblem}. \\

\textbf{Optimal $\rho$ setting.}  Again different values of $T$ gives different optimal values of $\rho$. However, the setting of $\alpha = \rho T^{-1}$, with a $\rho=\dfrac{1}{200}$ behaves quite well for all the three considered cases of $T=200, 1000, 5000$.\\

CoRNN \cite{rusch2020coupled} and IndRNN \cite{li2018independently} achieve the best performance in the literature for the AddProb task. They use 128 hidden units, so the same size has been implemented here for all of the the three vanilla RNN, LSTM, and roaRNN.
The AddProb turned out a less challenging task for both LSTMs and vanilla RNN. Their results are better than those usually reported in the literature \cite{arjovsky2016unitary, li2019deep, le2015simple, rotman2020shuffling}, probably due to the orthogonal initialisation scheme implemented here. However, for long sequences of $T=5000$, LSTM struggles to solve the task in reasonable time, and vanilla RNN already fails with $T=1000$. Contrariwise, roaRNN converges fast, within the first 50 hundred training steps in all the considered cases. Loss curves of Figure \ref{fig:addingproblem} are plotted in hundreds of training iterations in order to compare the results with others in the literature, among which \cite{rusch2020coupled} and \cite{li2018independently}.


\subsection{Permuted sequential MNIST task}
\label{sec:psMNIST}

\begin{figure*}[!t]
\centering
\begin{minipage}[c]{0.22\textwidth}
\centering
    \includegraphics[keepaspectratio=true,scale=0.33]{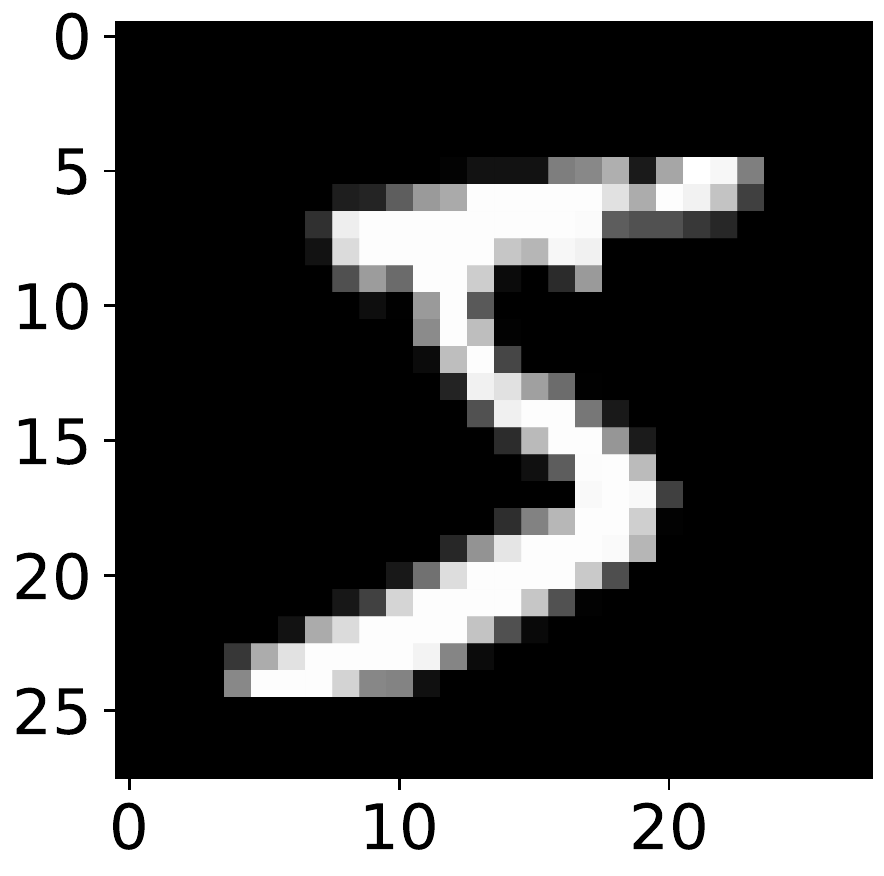}
\end{minipage}
\begin{minipage}[c]{0.7\textwidth}
    \centering
    \includegraphics[keepaspectratio=true,scale=0.24]{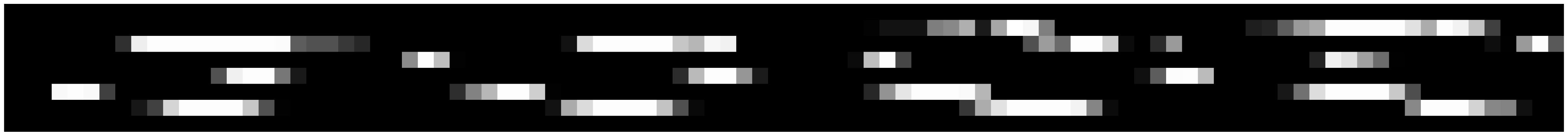}\\
    \vspace{0.35cm}
    \includegraphics[keepaspectratio=true,scale=0.24]{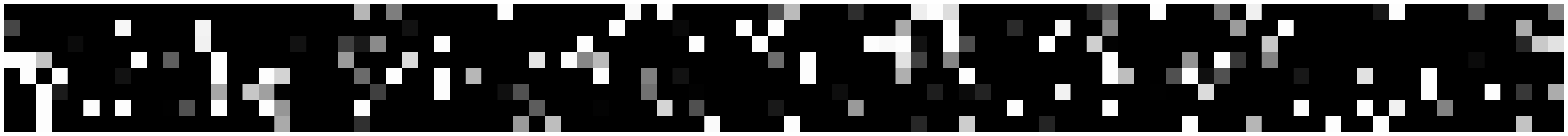}
\end{minipage}
\caption{\textbf{Left.} The first training sample of the MNIST dataset is a 28$\times$28 pixel image representing an handwritten digit 5. \textbf{Top right.} The 28$\times$28 pixel image representing the digit 5 is flattened in a $784=28^2$ sequence starting from the top left corner, proceeding from left to right and from top to bottom, until the bottom right corner. The sequence is plotted in 8 rows for a better visualisation. \textbf{Bottom right.} The resulting sequence is then permuted according to a predetermined permutation.}
\label{fig:fiveMNIST}
\end{figure*}

\textbf{Task.} Training dataset is the whole standard MNIST set of 60k samples. Test dataset is the whole standard MNIST set of 10k samples. Training and test input samples are flattened (from left to right and top to bottom) to be sequences of $T=784$ pixels. Then all the sequences are rearranged according to a given permutation, see Figure \ref{fig:fiveMNIST}. The permutation has been taken from https://www.nengo.ai/nengo-dl/examples/lmu.html, in order to faithfully compare the results obtained in this work with the results of LMU \cite{voelker2019legendre}, and NRU \cite{chandar2019towards}. \\

\textbf{Optimal $\rho$ setting.} Among the grid of $\rho$ values of $ \{ \frac{1}{300} , \frac{1}{200} , \frac{1}{100} , \frac{1}{50} , \frac{1}{10}, \frac{1}{2},  1, 2, 3, 5, 10, 50 \} $, the best results have been found with $\rho=\dfrac{1}{2}$, i.e. setting the hyperparameter $\alpha=\frac{1}{2 T}$.\\

\begin{figure}
    
    
        \resizebox{\linewidth}{!}{

            \begin{tabular}{ccc}
            \hline
            \textbf{Model} & \textbf{Parameters}  & \textbf{Test accuracy}\\
            \hline
            Vanilla RNN \cite{chang2017dilated} & $\approx$68k  & $ 71.6\%$\\
            LSTM \cite{helfrich2018orthogonal} & $\approx$270k  & $ 92.9\%$\\
            GRU \cite{chang2017dilated}  & $\approx$200k  & $ 94.1\%$\\
            Dilated CNN \cite{chang2017dilated}  & $\approx$46k  & $ 96.7\%$\\
            FC uRNN \cite{wisdom2016full} & $\approx$270k   & $ 94.1\%$\\
            BN LSTM \cite{cooijmans2016recurrent}  & $-$  &  $ 95.4\%$\\
            expRNN \cite{lezcano2019cheap}  & $\approx$137k  &  $ 96.6\%$\\
            Lipschitz RNN \cite{erichson2020lipschitz}  & $\approx$ 34k   &  $ 96.3\%$\\
            res-IndRNN \cite{li2019deep}  & $-$  &  $ 97.02\%$\\
            dense-IndRNN \cite{li2019deep}  & $-$   & $ 97.2\%$\\
            Shuffling RNN \cite{rotman2020shuffling}  & $\approx$50k  &  $ 96.43\%$\\
            NRU \cite{chandar2019towards}  & $\approx$165k  &  $ 95.38\%$\\
            LMU \cite{voelker2019legendre}   & $\approx$102k  & $ 97.15\%$\\
            coRNN \cite{rusch2020coupled}  & $\approx$134k  &  $ 97.3\%$\\
            \hline
            roaRNN  & $\approx$ 34k   & $ 97.24\%$\\
            roaRNN  & $\approx$ 69k   & $ 97.88\%$\\
            roaRNN  & $\approx$270k  & $ 98.25\%$\\
            \hline
            \end{tabular}
        }
        \label{tab:psMNIST}
        
    \vspace{0.5cm}
    
        \includegraphics[keepaspectratio=true,scale=0.38]{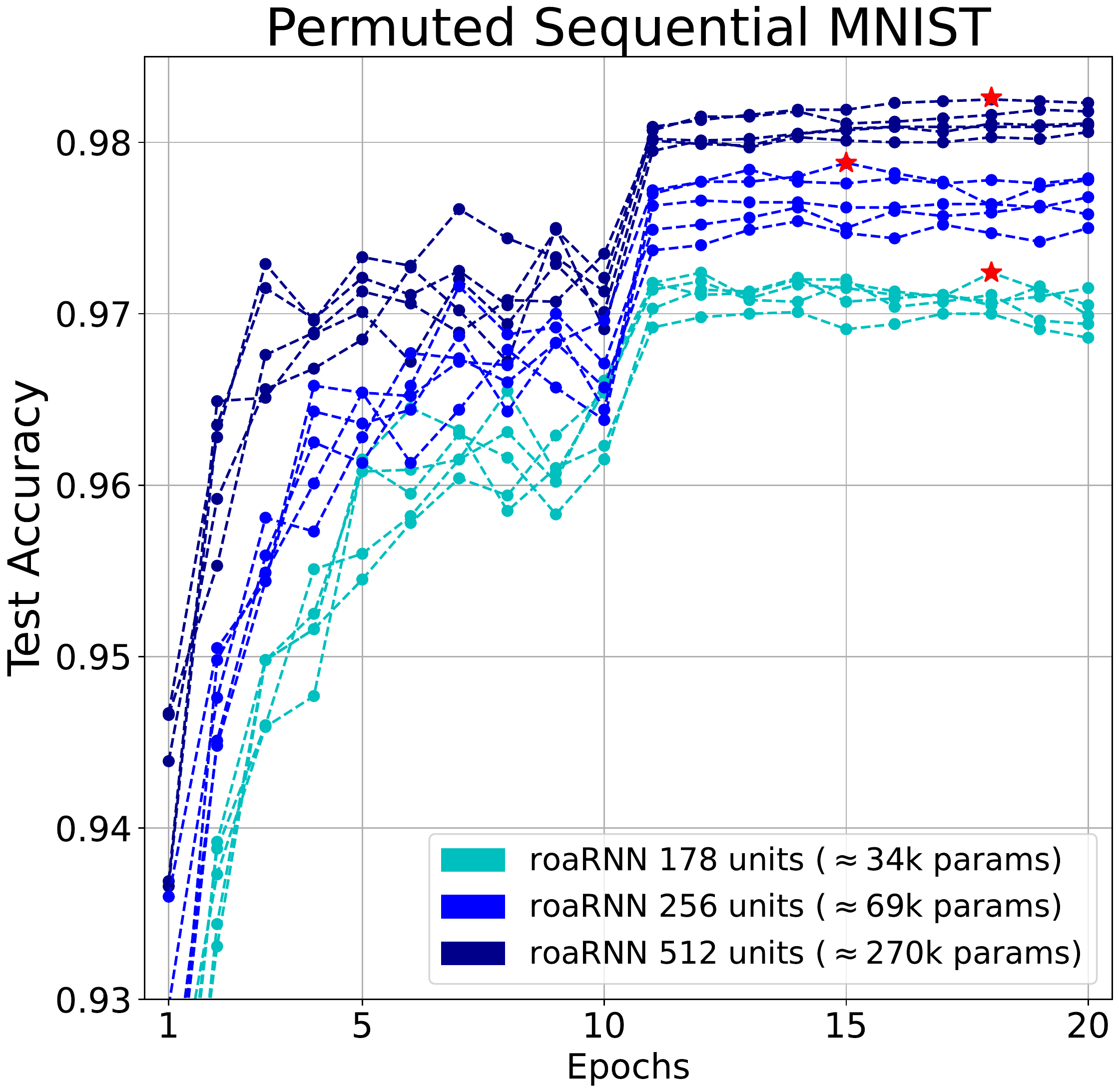}
          \caption{Test accuracy on psMNIST for several deep learning architectures compared to the performance of roaRNN of different sizes. Red stars mark the highest test accuracy reached in 20 epochs for each one of the three sizes.}%
          \label{fig:sequential_permuted_MNIST}

\end{figure}

In the table of Figure \ref{fig:sequential_permuted_MNIST} are reported the best test accuracy values on the psMNIST task reached by a wide plethora of neural network models ranging from stacked multilayer architectures (as dense-IndRNN and Dilated CNN), gated architectures (as GRU and LSTM), models provided with memory cells (as NRU and LMU), unitary learning models (as FC uRNN and expRNN), continuous-time ODE models (as Lipschitz RNN), second-order models (as coRNN), and other models specifically devised to tackle the V/E gradient issue of RNNs.
Each result reported in the table is accompanied with the reference in the literature where it is found. For some of them, the number of trainable parameters is not available.
However, note that these results are not fully fairly comparable to each other due to the different training method (optimiser, batch size, etc.), and permutation used for training the models.\\

RoaRNNs with hidden sizes of $ 178, 256$, and $512$, have been trained for a number of 5 trials each. The accuracy on the whole test dataset has been computed at the end of each epoch, and plotted in Figure \ref{fig:sequential_permuted_MNIST} for each trial. The darker the blue colour the larger the size of the recurrent layer. In all cases the output layer consists of 10 neurons, one for each digit to classify between $\{0, 1, \ldots, 9 \}$. The loss function is the cross-entropy. Training data is shuffled at the end of each epoch. Batch size is 100. Learning rate is $0.1$ for the first 10 epochs, and then dropped to $0.01$ from the 11th epoch.

Remarkably, most of the neural network architectures reported in the table need to be trained for several dozens (if not hundreds) of epochs to reach good performance. On the contrary, roaRNN reaches test-accuracy values above $97\%$ after just 10 epochs in all cases, outperforming rapidly the vast majority of alternative deep learning architectures.

\subsection{Ablation study: identity vs random orthogonal matrices}
\label{sec:ortho_role}

\begin{figure}[ht!]
    \centering
    \includegraphics[keepaspectratio=true,scale=0.12]{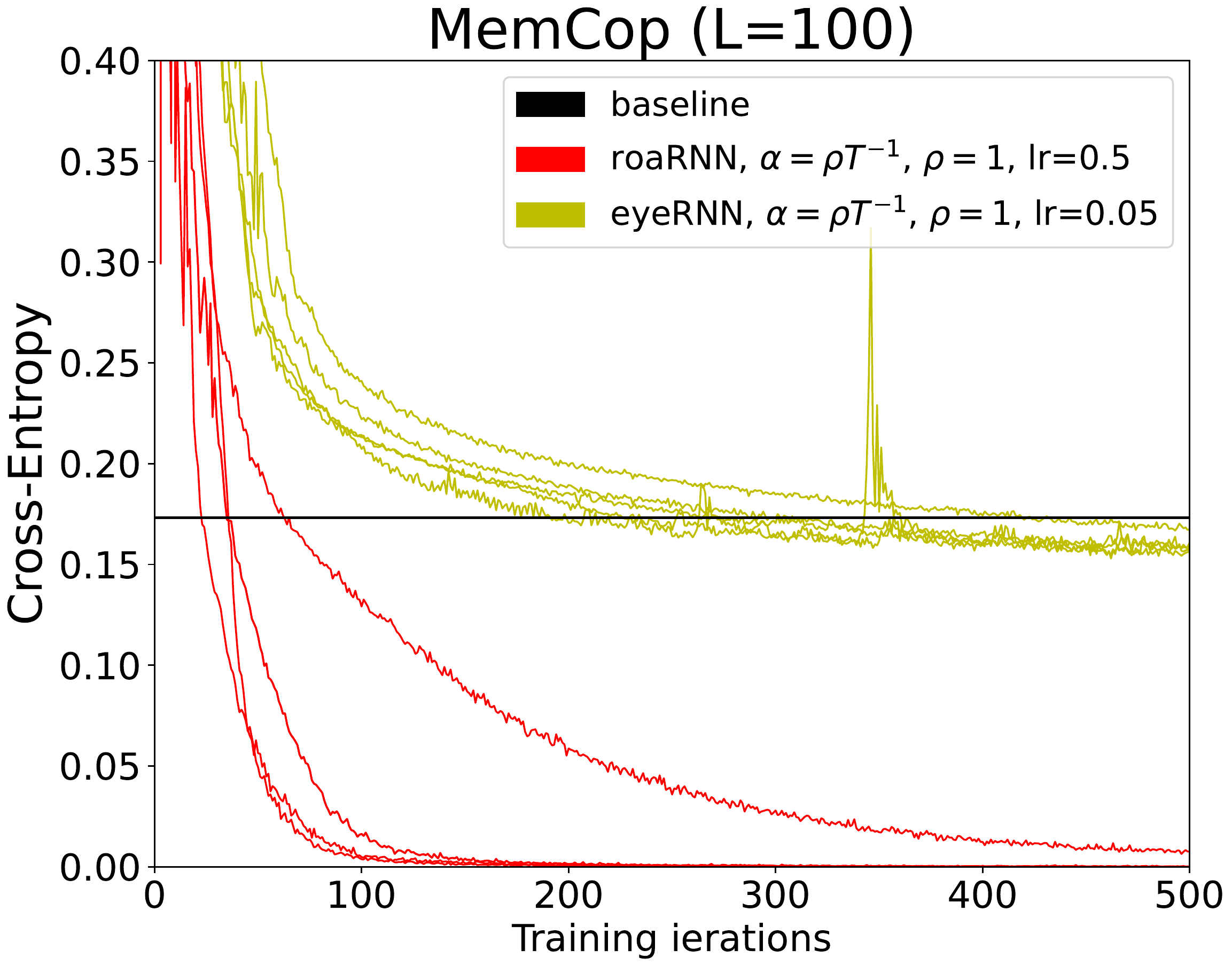}~\includegraphics[keepaspectratio=true,scale=0.12]{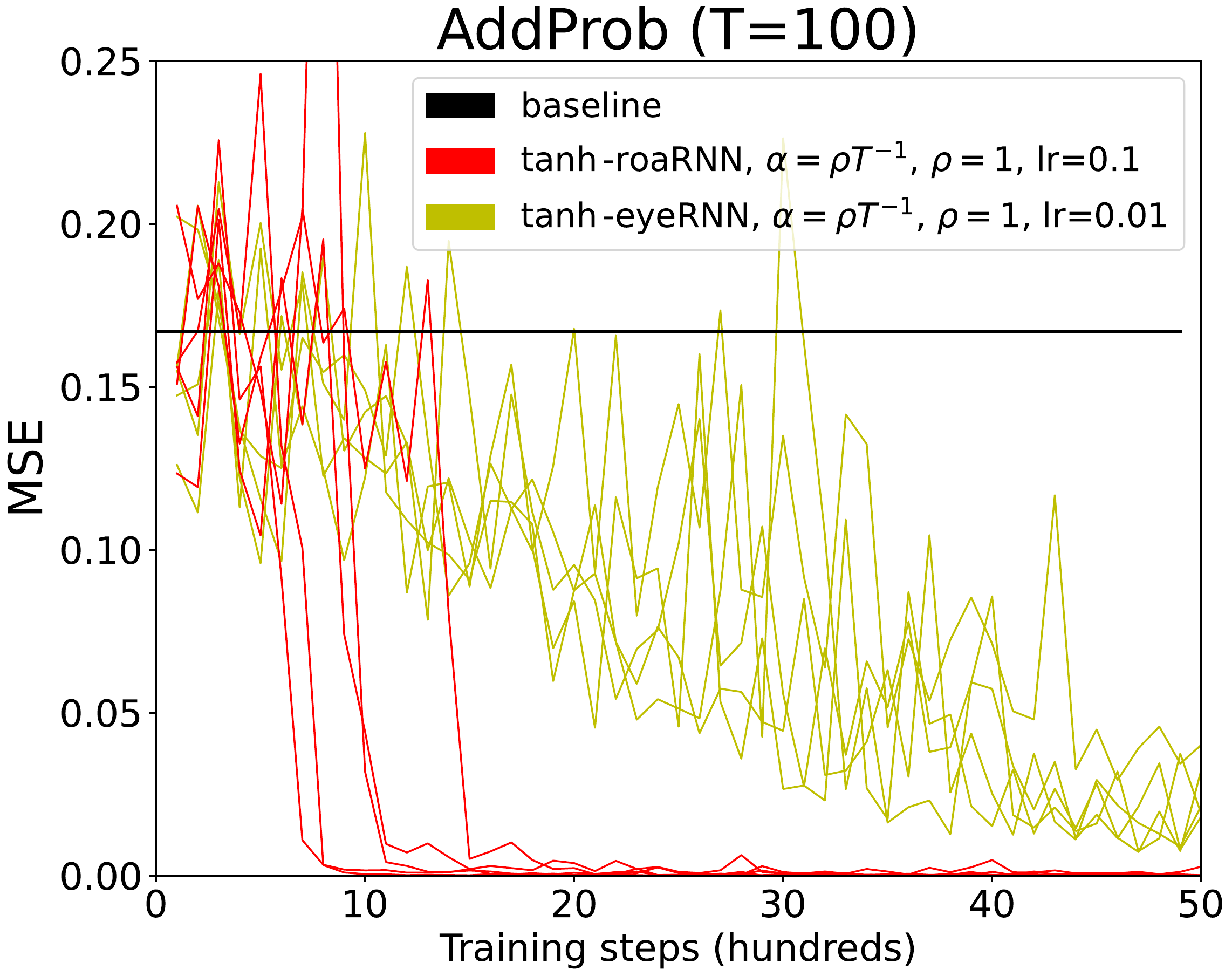}~\includegraphics[keepaspectratio=true,scale=0.12]{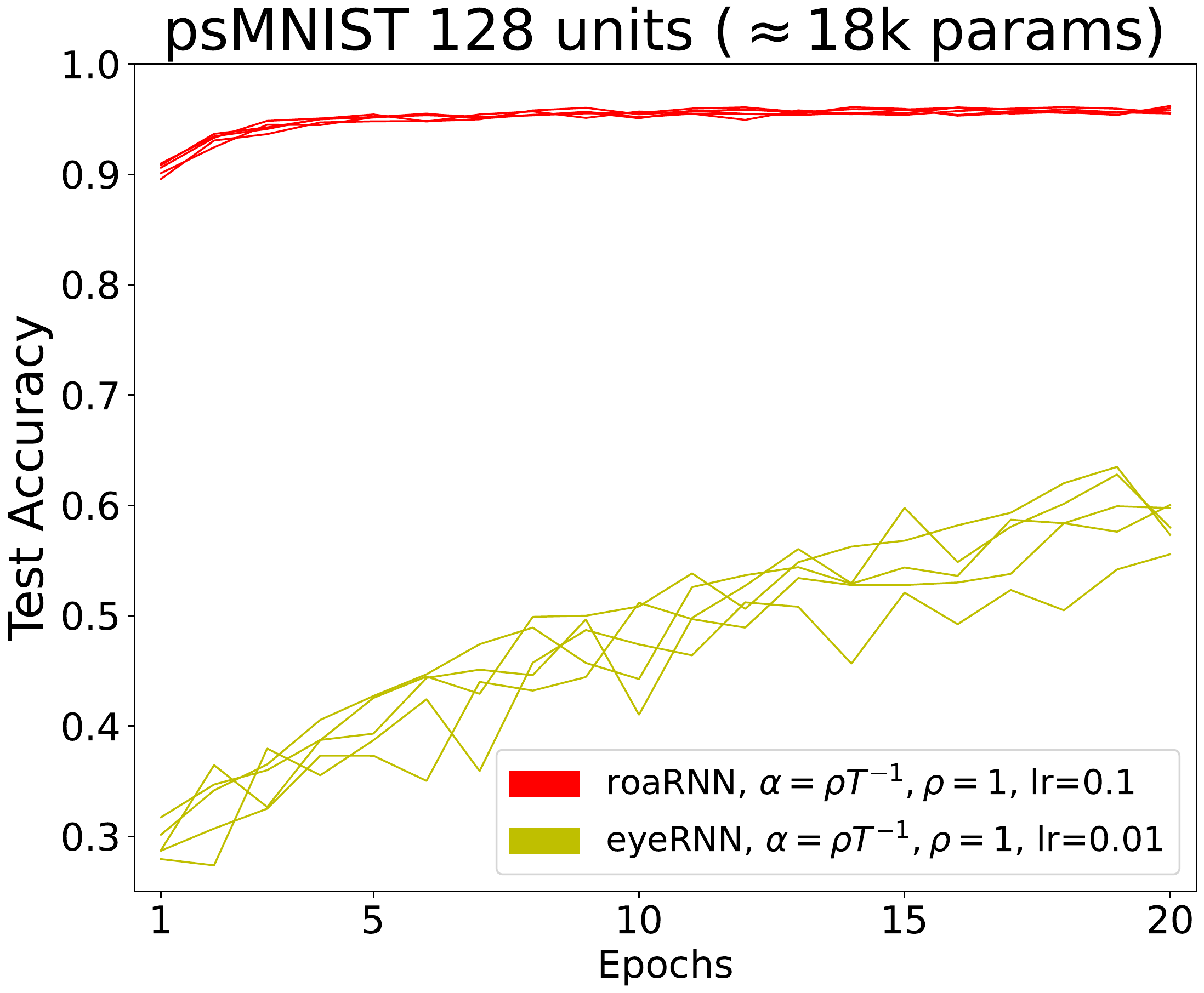}
\caption{Comparison between the recurrent neural network model of eq.s \eqref{eq:ortho_hiddenRNN}-\eqref{eq:ortho_outputRNN} where the orthogonal matrix $O$ is the identity matrix (eyeRNN), and a randomly generated one (roaRNN).}
\label{fig:id_vs_ortho}
\end{figure}

In this section, the performance of the RNN model of eq.s \eqref{eq:ortho_hiddenRNN}-\eqref{eq:ortho_outputRNN} with random orthogonal $O$ (roaRNN) are compared against the peculiar case of the identity matrix $O=I$ (eyeRNN), see Figure \ref{fig:id_vs_ortho}.
Specifically for both models, 5 simulations based on 5 different initialisation of parameters have been run, for each one of the three selected tasks (MemCop, AddProb, psMNIST). 
In particular, the roaRNN's and eyeRNN's parameters are made to exactly coincide when learning starts\footnote{This is ensured by using the same set of 5 seeds for each task.}, i.e. differing exclusively between each other in the choice of the matrix $O$. 
Preliminary trials on eyeRNN (not reported here) revealed that a good learning rate for eyeRNN is generally ten times smaller than the one for roaRNN.
Both models have been run with 128 hidden units, Adam optimiser, and a standard choice of $\rho =1 $, in all the three tasks.
In the AddProb task, eyeRNN performed significantly better with $\tanh$ activation rather than ReLU. Therefore, for this task the comparison has been done between $\tanh$-eyeRNN and $\tanh$-roaRNN. While, MemCop and psMNIST have been run with ReLU.
For the psMNIST task, training was stopped at the 20th epoch, albeit eyeRNN models were still learning, since it was enough to prove the disparity between the two variants.

Note that the bounds on the distribution of singular values of the IOJ provided in Theorem \ref{thm:distribution_singvals} hold for any orthogonal matrix, hence for the identity matrix as well. Nevertheless, as it is suggested by the results in Figure \ref{fig:id_vs_ortho}, the identity does not give results as good as a random orthogonal matrix does.
\emph{The conjecture is that 
the particular annular distribution of eigenvalues of the roaRNN gradient helps the learning dynamics to reach a desired eigenspectrum configuration for the task at hand. While for the case of the identity matrix, the eigenspectrum of the RNN gradient is not heterogeneously spread over the unitary circle, and thus it might need a greater learning effort to adapt itself on certain tasks.}
An extended analysis would be ideal to establish which are the important features that an orthogonal matrix must possess to facilitate further the learning process.

\section{Conclusion}
In this paper a new technique to pursue approximate dynamical isometry in generic nonlinear FNNs is proposed.
Loosely speaking, it suffices to add fixed random semi-orthogonal matrices between layers, suitably weighted via convex combination with the nonlinear trainable residual component, to consent the signals to flow back and forth along the NN. 
The same intuition leads to a slight variation of the vanilla RNN model, called roaRNN, which can be trained via backpropagation through-time to learn arbitrarily long-term dependencies in the input.
The performance of a roaRNN is exceptional considered its ease of implementation
, and at the cost of just one more hyperparameter to tune, i.e. $\rho$.

The simplicity of the proposed model makes it amenable to a rigorous mathematical analysis of the gradient update dynamics, and yields an intelligible solution to a problem that afflicted neural network models from the early nineties.
Nevertheless, many aspects still need to be further explored. 

For example, although locally to each parameter's update the gradient dynamics of the proposed model have been proved in this paper to be stable, a theoretical investigation on the global learning dynamics has been neglected.
The experiments run on both roaFNN and roaRNN indicate that a favourable side effect of exploiting random orthogonal ``filters" might be to effectively regularise the learning dynamics. 
How these orthogonal matrices alter the geometry of the loss landscape is a compelling research question. 

Moreover, further investigations are needed to enlighten the role that specific structured orthogonal matrices can have on the proposed model, as suggested by the difference of performance observed between the case of the identity matrix versus random orthogonal matrices.


Besides, an implicit assumption of the proposed model is the existence of an optimum scalar value for the hyperparameter $\rho$. However, there might be complex tasks where the nonlinear dependence in $\rho$ is intrinsically nonconvex, resulting in many local optimal values; or cases where it might be helpful to retrieve information in the past activations at different time lags for different spatial components. 
This makes interesting to experiment, albeit more computationally expensive, an adapting multidimensional vector $\rho$ in the proposed model.

Finally, although many of the proposed solutions in the literature so far to combat V/E gradients (e.g. skipping connections, batch normalisation, trainable gates, etc.) could still be implemented on the architecture proposed here, it should be matter of research to establish whether the combination of these techniques comports more advantages than unreasonable complexity in a model which is already afflicted by interpretability and explainability issues.

\section*{Acknowledgments}
I would like to thank personally Lorenzo Livi, Peter Ashwin, Claudio Gallicchio for their insightful comments and intellectual guidance, and the Mathematical Sciences EPSCR (University of Exeter) and MyBreathingHeart Project (University of Pisa) whose funds made this work possible.


{\appendices

\section{Bounds on the maximum singular value of the IOJ of a roaFNN}
\label{sec:bounds}

Recall the roaFNN's equation
\begin{equation*}
    x_{l+1}  = \alpha\phi( W_l x_l + b_l ) \,\, + \,\, (1-\alpha) O_l x_l, \qquad l=0,1, \ldots , L-1.
\end{equation*}
Recall that, for $l=0, \ldots, L-1$, matrices $W_l $ have dimensions $ N_{l+1}\times N_l $, and $ O_l $ are random semi-orthogonal matrices of the same dimensions. 
Recall also that $ d_l = \phi'(W_l x_l + b_l) $ are vectors that vary in $[0,r]^{N_{l+1}}$, and denote $\sigma_l :=  \lVert W_l \rVert $, and $\sigma_{max} := \max_l \sigma_l $, where $\lVert \cdot \rVert $ denotes the matrix norm induced by the Euclidean norm of vectors, i.e. the maximum singular value.
Let us define matrices 
$$
P_{l}  := P_{l+1} \bigl[ \alpha \,\, \diag(d_l) \,\, W_l \,\, + \,\, (1-\alpha)\,\, O_l \bigl],
$$
starting with $P_{L}$ as the identity matrix of dimension $N_L$. 
Thus, by definition $ P_1 $ is the IOJ of the roaFNN.\\


\begin{lem}
\label{lem:upper_recurs}
For each $ l=1, \ldots, L-1 $, the following inequality holds
$
     \lVert P_{l} \rVert \leq \bigl[ 1 + \alpha(r\sigma_l - 1) \bigl] \lVert P_{l+1} \rVert .
$
\end{lem}
\proof
It follows from the triangle inequality applied on the matrix $P_{l}$. In fact,
$
    \label{eq:triangle}
    \lVert  P_{l} \rVert
    = 
    \lVert \alpha  P_{l+1} \diag(d_l)  W_l \,\,   + \,\, (1-\alpha) P_{l+1}  O_l \rVert 
    \leq
    \alpha \lVert P_{l+1}  \diag(d_l)  W_l \rVert + (1-\alpha) \lVert P_{l+1}  O_l \rVert 
     \leq
    \alpha r\sigma_l \lVert  P_{l+1} \rVert + (1-\alpha)\,\, \lVert  P_{l+1} \rVert
    = \bigl[ 1 + \alpha(r\sigma_l - 1) \bigl] \lVert P_{l+1} \rVert.
$
The second inequality follows from the fact that $d_l \in [0,r]^{N_{l+1}}$, the definition of $\sigma_l =  \lVert W_l \rVert $, and the isometric property of semi-orthogonal matrices which implies $\lVert O_{l} \rVert = 1 $.
$\qed$\\


%
\begin{lem}
\label{lem:same_norm}
Let be given a matrix $P$ of dimension $h\times m $, and a semi-orthogonal matrix $O$ of dimension $ m\times n $.
If $m\leq n$, then the following holds
$
    \lVert P O \rVert =  \lVert P  \rVert .
$
\end{lem}
\proof
First of all, by submultiplicativity, and the isometric property of $O$, holds $ \lVert P O \rVert \leq  \lVert P  \rVert  \lVert O  \rVert \leq  \lVert P  \rVert$.
By contradiction, assume that $ \lVert P O \rVert <  \lVert P  \rVert$.
This must be true for the transpose of these matrices as well. Hence we have that $ \|  O^T P^T \| < \| P^T \| $.
Let us denote $ v_0, v_1 \in \mathbb{R}^{h}$, the two unitary vectors that maximise, respectively, the linear operators $ O^T P^T $, and $ P^T $. 
By assumption of $ \lVert O^T P^T \rVert <  \lVert  P^T  \rVert$ we have that
\begin{equation}
    \label{eq:absurd}
    \| O^T P^T v_0 \| < \| P^T v_1 \|.
\end{equation}
By hypothesis $O$ is a semi-orthogonal matrix of dimension $m \times n$ with $ m \leq n $. This implies that $ O O^T = I_m $, which in turn implies that $ \| O^T w \| = \| w \|  $, for all $w \in \mathbb{R}^{m} $. In fact, $ \| O^T w \|^2 =  (O^T w)^T  O^T w = w^T O O^T w = w^T w = \| w \|^2 $.
Therefore, for the vector $ w_1 = P^T v_1$ we get 
\begin{equation}
    \label{eq:contradiction}
    \|  P^T v_1 \|  = \|  w_1 \| = \| O^T  w_1 \| = \| O^T P^T v_1 \| \leq \| O^T P^T v_0 \|,
\end{equation}
where the last inequality holds since $v_0$ maximises the linear operator $O^T P^T$. The inequality \eqref{eq:contradiction} contradicts \eqref{eq:absurd}.
$\qed$\\


%
\begin{lem}
\label{lem:recurs_bound}
Let us assume a roaFNN model with non-increasing sizes of layers, i.e. assume $ N_0 \geq N_1 \geq \ldots, \geq N_L $.  
If $\alpha < \dfrac{1}{1+r\sigma_{\text{max}}} $, then it holds
$
    \lVert P_{l} \rVert \geq \bigl[ 1 - \alpha(1+r\sigma_l) \bigl] \lVert P_{l+1} \rVert , 
$
for $ l=1, \ldots, L-1$.
\end{lem}
\proof
The hypothesis that $\alpha < \dfrac{1}{1+r\sigma_{\text{max}}} $ implies that $ r\sigma_l \alpha < 1 - \alpha $, for all $ l=0, \ldots, L-1$. Thus, it follows that 
\begin{equation}
    \label{eq:first_ineq}
    \alpha \lVert P_{l+1} \diag(d_l) W_l  \rVert 
    \leq 
    \alpha \sigma_l r \lVert P_{l+1} \rVert
    \leq 
    (1-\alpha) \lVert P_{l+1} \rVert .
\end{equation}
Now, by hypothesis $ N_{l+1}\leq N_l $, so Lemma \ref{lem:same_norm} ensures that $ \| P_{l+1} \| = \| P_{l+1} O_l \|  $. Hence, from \eqref{eq:first_ineq} we obtain 
\begin{equation}
    \label{eq:first_ineq_samenorm}
    \alpha \| P_{l+1} \diag(d_l) W_l\| 
    \leq 
    (1-\alpha) \| P_{l+1} O_l \|.
\end{equation}
Therefore, \eqref{eq:first_ineq_samenorm} allows us to write the reverse triangle inequality in the following form
$
    (1-\alpha)\,\, \lVert  P_{l+1} O_l  \rVert - \alpha \lVert P_{l+1} \diag(d_l)  W_l    \rVert
\leq
\lVert \alpha P_{l+1} \diag(d_l)  W_l  \,\, + \,\, (1-\alpha) P_{l+1}  O_l  \rVert = \lVert  P_{l} \rVert,
$
which can be rewritten, according to Lemma \ref{lem:same_norm}, as 
$
    \label{eq:second_ineq}
    (1-\alpha)\,\, \lVert  P_{l+1}  \rVert - \alpha \lVert P_{l+1} \diag(d_l)  W_l    \rVert
    \leq \lVert  P_{l} \rVert.
$
The left term of the last inequality can be bounded from the bottom using the first inequality in \eqref{eq:first_ineq} obtaining
\begin{equation}
    \label{eq:third_ineq}
    (1-\alpha)\,\, \lVert  P_{l+1}  \rVert - \alpha \sigma_l r \lVert P_{l+1} \rVert 
    \leq \lVert  P_{l} \rVert,
\end{equation}
which can be rearranged as $ \bigl[ 1 - \alpha(1+r\sigma_l) \bigl] \| P_{l+1} \| \leq \| P_l \| $, and that is the thesis.
$\qed$\\

\begin{thm}
\label{thm:_main_thm}
Let us consider a roaFNN model with an activation function $\phi$ such that its derivative assumes values in $[0,r]$.
Let us denote $\sigma_{\text{max}} := \max_l \lVert W_l \rVert$. 
If we parametrise $\alpha = \rho (L-1)^{-1}$, with a positive real valued $\rho$, then the maximum singular value of the IOJ of the roaFNN admits the following $L$-independent upper bound
$
    \biggl\lVert \dfrac{\partial x_{L}}{ \partial x_{1}} \biggl\rVert \,\, \leq \,\, \exp\bigl( \rho(r\sigma_{\text{max}} - 1) \bigl).
$
Moreover, if $\rho < \dfrac{L-1}{1+r\sigma_{\text{max}}}$, and assuming non-increasing sizes of layers, then it holds the following $L$-independent lower bound
$
    \exp\bigl(-\rho(1+r\sigma_{\text{max}})\bigl) \,\, \leq \,\, \biggl\lVert  \dfrac{\partial x_{L}}{ \partial x_{1}} \biggl\rVert.
$
\end{thm}
\proof
By definition $ P_1 = 
 \bigl[ \alpha  \diag(d_{L-1}) W_{L-1}  +  (1-\alpha) O_{L-1} \bigl]  \ldots \bigl[ \alpha  \diag(d_{1}) W_{1}  +  (1-\alpha) O_{1} \bigl]
=
\dfrac{\partial x_{L}}{\partial x_1}  $ is the IOJ of the roaFNN model.
\textbf{Upper bound.} Lemma \ref{lem:upper_recurs} implies that
$\lVert P_1 \rVert \leq \prod_{l=1}^{L-1}  1 + \alpha(r\sigma_l-1)$, which can be upper-bounded using the definition of $\sigma_{\text{max}} = \max_l \sigma_l $ by the term $ \bigl[ 1 + \alpha(r \sigma_{\text{max}}-1) \bigl]^{L-1}$. Now, with the parametrisation $\alpha = \rho (L-1)^{-1}$, it holds
$
    \biggl\lVert \dfrac{\partial x_{L}}{\partial x_1} \biggl\rVert   
    \leq
    \bigl[  1 + \dfrac{\rho (r\sigma_{\text{max}}-1)}{L-1}  \bigl]^{L-1} 
    \leq 
    \me^{\rho (r\sigma_{\text{max}}-1) },
$
where the last inequality holds since $ ( 1+ \dfrac{A}{x})^x < \me^A   $ for all $x>0$.
\textbf{Lower bound.} Lemma \ref{lem:recurs_bound} implies that
$\lVert P_1 \rVert \geq \prod_{l=1}^{L-1}  1 - \alpha(1+r\sigma_l)$, which can be lower-bounded using the definition of $\sigma_{\text{max}} = \max_l \sigma_l $ by the term $ \bigl[ 1 - \alpha(1+r \sigma_{\text{max}}) \bigl]^{L-1}$. Now, with the parametrisation $\alpha = \rho (L-1)^{-1}$, it holds
\begin{align*}
    \biggl\lVert \dfrac{\partial x_{L}}{\partial x_1} \biggl\rVert 
    &\geq  
    \bigl[  1 - \dfrac{\rho(1+r\sigma_{\text{max}})}{L-1} \bigl]^{L-1} \geq\\
    &\geq
    \bigl[  1 - \dfrac{\rho(1+r\sigma_{\text{max}})}{L-1} \bigl]^{L}   
    \geq 
    \me^{ - \rho(1+r\sigma_{\text{max}}) },
\end{align*}
where the second inequality holds because of the hypothesis that $\rho < \dfrac{L-1}{1+r\sigma_{\text{max}}}$, and the last inequality holds because $  ( 1+  \dfrac{A}{x})^{x+1} >  \me^A $ for all $x>0$. 
$\qed$

\section{A lower bound for the minimum singular value of the IOJ of a roaRNN}
\label{sec:minimum_sing_val}

In the case of roaRNN we deal with square matrices. If they are invertible then we can easily find a lower bound for the minimum singular value of the IOJ matrix $J$, just exploiting the fact that
$\sigma_{min}(J) = \dfrac{1}{\lVert J^{-1} \rVert} $.\\


\begin{lem}
\label{lem:inverse_sum}
Let be given two square matrices $A, B \in \mathbb{R}^{N \times N}$ such that $A$ is invertible and $A^{-1}B$ has spectral radius less than 1. Then it holds the following relation
$
    (A-B)^{-1} = \biggl[\sum_{k=0}^{\infty} (A^{-1}B)^k \biggl] A^{-1}.
$
\end{lem}
\proof
The formal geometric series $ \sum_{k=0}^{\infty} C^k = (I-C)^{-1} $ holds for all squared matrices $C$ having spectral radius less than 1. Therefore, by hypothesis it holds
$
\sum_{k=0}^{\infty} (A^{-1}B)^k = (I-A^{-1}B)^{-1}.
$
Therefore, multiplying on the right both terms of the above equation by the matrix $A^{-1}$ we get
$
    \biggl[\sum_{k=0}^{\infty} (A^{-1}B)^k \biggl] A^{-1} = (I-A^{-1}B)^{-1} A^{-1} 
    = \bigl(A (I-A^{-1}B)\bigl)^{-1} = (A-B)^{-1},
$
which is the thesis.
$\qed$\\


\begin{lem}
\label{lem:invertibility}
Let be given a matrix $G$ with $ \gamma =\lVert G \rVert$, and an orthogonal matrix $O$. If $ 0<\alpha < \dfrac{1}{1+ \gamma}$, then the matrix $\alpha W + (1-\alpha)O $ is invertible.
\end{lem}
\proof
I will prove that the kernel of $ \alpha G + (1-\alpha)O $ is null. 
Assume there exists a vector $v \neq 0 $ such that $ \bigl[\alpha G + (1-\alpha)O\bigl]v = 0 $. Then, it follows that $ (1-\alpha)O v = -\alpha G v  $. Now, by hypothesis $\alpha \gamma < (1-\alpha) $, hence we have that $ \lVert -\alpha G v \rVert \leq \alpha \gamma \lVert  v \rVert <  (1-\alpha) \lVert v \rVert = \lVert (1-\alpha)O v \rVert $. This contradicts the assumption $  (1-\alpha)O v = -\alpha G v  $.
$\qed$\\


\begin{lem}
\label{lem:upper_bound_inverse}
Consider model \eqref{eq:ortho_hiddenRNN} with activation function $\phi$ such that its derivative is bounded in $[0,r]$, and parameterised via $ \alpha = \rho (L-1)^{-1}$.
Denote $\sigma=\lVert W_h \rVert $. If $ \rho < \dfrac{L-1}{1+r \sigma} $, then for each $l=0, \ldots, L-2$, the matrix $ \dfrac{\partial x_{l+1}}{\partial x_l} = \alpha  \diag(\phi'(y_{l}))  W_{h}   +  (1-\alpha)  O $ is invertible and its inverse admits the following upper bound
$
\biggl\lVert \Bigl(\dfrac{\partial x_{l+1}}{\partial x_l} \Bigl)^{-1} \biggl\rVert \leq \dfrac{1}{1-\alpha(1+ r \sigma) }.  
$
\end{lem}
\proof
I will use the expression of Lemma \ref{lem:inverse_sum}, $(A-B)^{-1} =\sum_{k=0}^{\infty} (A^{-1}B)^k A^{-1} $, with $B=-\alpha \diag(\phi'(y_{l})) W_h $, and $A=(1-\alpha)O $, for which I know that $A^{-1}=(1-\alpha)^{-1}O^{-1}$. Therefore, by the triangle inequality I know that $ \lVert \bigl(\dfrac{\partial x_{l+1}}{\partial x_l} \bigl)^{-1} \rVert = \lVert \sum_{k=0}^{\infty} (A^{-1}B)^k A^{-1} \rVert \leq \sum_{k=0}^{\infty} \lVert (A^{-1}B)^k A^{-1} \rVert $. Then by submultiplicativity holds $ \sum_{k=0}^{\infty} \lVert (A^{-1}B)^k A^{-1} \rVert \leq \sum_{k=0}^{\infty} \lVert (A^{-1}B)^k \rVert \lVert A^{-1} \rVert $. Again submultiplicativity implies $ \lVert (A^{-1}B)^k \rVert \leq \lVert (A^{-1}B) \rVert ^k \leq \lVert A^{-1} \rVert ^k \lVert B \rVert ^k = \dfrac{1}{(1-\alpha)^k} \lVert B \rVert ^k \leq \dfrac{1}{(1-\alpha)^k} (\alpha r \sigma)^k  $. All together we have the bound 
$
\lVert \bigl(\dfrac{\partial x_{l+1}}{\partial x_l} \bigl)^{-1} \rVert \leq (1-\alpha)^{-1} \sum_{k=0}^{\infty} (\dfrac{\alpha r \sigma}{1-\alpha} )^k .
$
The right term of the above inequality converges to 
$
\dfrac{1}{1-\alpha(1+ r \sigma) },
$
as long as $\dfrac{\alpha r \sigma}{1-\alpha}<1$, that is when $ \alpha(1+ r \sigma) < 1 $. Assuming $\alpha=\rho(L-1)^{-1}$ the condition for the convergence is equivalent to $ \rho < \dfrac{L-1}{1+r \sigma}$.
$\qed$\\


\begin{thm}
Let us consider a roaRNN \eqref{eq:ortho_hiddenRNN} with an activation function $\phi$ such that its derivative assumes values in $[0,r]$, and let us denote $\sigma=  \lVert W_h \rVert$.
If we parametrise $\alpha = \rho (L-1)^{-1}$, with a positive real valued $\rho$ such that $\rho < \dfrac{L-1}{1+r\sigma}$, then matrices $\dfrac{\partial x_L }{ \partial x_{s+1}}$ are invertible for each $s=0, \ldots, L-2$, and it holds the following $L$-independent lower bound for the minimum singular value of the IOJ 
$
    \me^{-\rho(1+ r\sigma)} \leq \sigma_{min}\Bigl(\dfrac{\partial x_L}{\partial x_1} \Bigl).
$
\end{thm}
\proof
First of all, it is known that for any invertible matrix $J$ it holds that $ \sigma_{min}(J) = \dfrac{1}{\lVert J^{-1} \rVert} $. In our case, we deal with $J=\dfrac{\partial x_L}{\partial x_1}$. I will find an upper bound for $ \lVert J^{-1} \rVert $, which in turns will lead to a lower bound for $ \sigma_{min}(J) $.
By definition $J$ is the product
$
    \bigl[ \alpha  \diag(\phi'(y_{L-1}))  W_{h}   +  (1-\alpha)  O \bigl]
    \ldots \bigl[ \alpha  \diag(\phi'(y_1))  W_{h}   +  (1-\alpha)  O \bigl].
$
Note that by Lemma \ref{lem:invertibility} we have that $J$ is a product of invertible matrices, thus an invertible matrix itself.
Now, 
$
\lVert J^{-1} \rVert 
= 
\biggl\lVert \Bigl( \prod_{l=L-1}^{1} \dfrac{\partial x_{l+1}}{\partial x_l} \Bigl)^{-1} \biggl\rVert
= 
\biggl\lVert  \prod_{l=1}^{L-1} \Bigl(\dfrac{\partial x_{l+1}}{\partial x_l} \Bigl)^{-1} \biggl\rVert
\leq 
 \prod_{l=1}^{L-1} \biggl\lVert \Bigl(\dfrac{\partial x_{l+1}}{\partial x_l} \Bigl)^{-1} \biggl\rVert 
 \leq 
  \dfrac{1}{\bigl[1-\alpha(1+ r \sigma)\bigl]^{L-1} } 
  \leq 
  \me^{\rho(1+r \sigma)}.
$
where the first inequality is by submultiplicativity, the second inequality is Lemma \ref{lem:upper_bound_inverse}, and the last is a known inequality of the exponential. Finally, we get the thesis
$
\sigma_{min}(J) = \dfrac{1}{\lVert J^{-1} \rVert} \geq \exp(-\rho(1+r \sigma)).
$
$\qed$

\section{Reducing memory and time consumption}
\label{sec:reducing_memory_time}

The model \eqref{eq:proposed_model} has now two times more real numbers to store than a standard multilayer perceptron. 
Two strategies to alleviate this problem are proposed:
\begin{itemize}
    \item[(I)] reducing memory and computational time consumption replacing random semi-orthogonal matrices with random \emph{semi-permutation matrices}. See def. \ref{def:permut} below for the definition of a semi-permutation matrix. From preliminary experiments, randomly generated semi-permutation matrices seem to work fine.
    \item[(II)] recycling the same semi-orthogonal matrix whenever the dimensions permit it. For instance, in a FNN where all the layers from input to output have the same dimension, the same orthogonal matrix might be exploited between all the layers. 
\end{itemize}

\begin{defn}
\label{def:permut}
A matrix $P \in \bR^{N\times M}$ is called \emph{semi-permutation} if either $     P = 
\begin{bmatrix}
    e_{\sigma(1)} , e_{\sigma(2)} , \ldots , e_{\sigma(M)}
\end{bmatrix} $ or $     P = 
\begin{bmatrix}
    e_{\sigma(1)} , e_{\sigma(2)} , \ldots ,
    e_{\sigma(N)} 
\end{bmatrix}^T $,
where $\sigma$ represents a permutation of either $M$ elements (if $N>M$) or $N$ elements (if $N\leq M$), and $e_{\sigma(j)}$ is the $\sigma(j)$-th column of the identity matrix of dimension either $M$ or N.
\end{defn}

\begin{remark}
Among all semi-orthogonal matrices, semi-permutation matrices are those that minimise the storage and computation resources. 
\end{remark}

\section{Continuous-time roaRNN and link to LipschitzRNNs}
\label{sec:CTrandorthoRNN}

The RNN model \eqref{eq:ortho_hiddenRNN} presented in this paper is discrete-time. 
Nevertheless, it can be derived from the following continuous-time RNN:
\begin{equation}
    \label{eq:CTorthoRNN}
    \tau \dot{x} = A x + \phi ( W_h x + b_h + W_i u  ) ,
\end{equation}
where $\tau$ can be viewed as a vector of time constants characterising the neuronal dynamics, and
\begin{equation}
    \label{eq:hidden_friction}
    A= \alpha^{-1}\bigl[ (1-\alpha)O - I \bigl],
\end{equation}
where $I$ is the identity matrix.
Integrating via the explicit Euler method with time step $\Delta t = \dfrac{1}{L-1} $, equation \eqref{eq:ortho_hiddenRNN} can be derived from \eqref{eq:CTorthoRNN} defining the time constant as $ \tau = \rho^{-1} $:
\begin{align*}
    x_{k+1} & = \alpha \phi( W_h x_k + b_h + W_i u_{k+1} ) + (1-\alpha)O x_k + x_k - x_k\\
    x_{k+1} & =  x_k + \Delta t \Biggl[ \dfrac{1}{\Delta t}\Bigl( \alpha \phi( W_h x_k + b_h + W_i u_{k+1} ) + \alpha A x_k   \Bigl) \Biggl] \\
    x_{k+1} & = x_k + \Delta t \Biggl[ \tau^{-1}\Bigl(  \phi( W_h x_k + b_h + W_i u_{k+1} ) + A x_k   \Bigl) \Biggl].
\end{align*}

\begin{remark}
The parameter $\alpha$ of a roaRNN can be viewed as a constant update gate (GRUs) or as a weight for the residual connection (ResNets, FastRNNs) or as a leaky parameter (leakyESNs).
Nevertheless, an alternative and more biologically-based interpretation is given by the continuous-time perspective of model \eqref{eq:CTorthoRNN}, where the vector $\alpha = \tau^{-1} \Delta t$ is directly proportional to the vector $\tau^{-1}$, i.e. the vector of the frequencies characterising the neurons of the recurrent network.
\end{remark}

The continuous-time model \eqref{eq:CTorthoRNN} shares the same equation with the LipschitzRNN \cite{erichson2020lipschitz}, a recently introduced RNN model.
There the authors propose to parametrise the hidden-to-hidden matrices of \eqref{eq:CTorthoRNN}, i.e. $A$ and $W_h$, as
$$
 (1-\beta)(M + M^T) + \beta (M-M^T) - \gamma I,
$$
with $M$ a trainable matrix, and two pairs of tunable parameters $\beta  \in [0.5,1], \gamma \geq 0 $, one for each of the two recurrent matrices $A$ and $W_h$.
On the other hand, in this paper the matrix $A$ is not optimised, it has the form of \eqref{eq:hidden_friction} with a randomly generated orthogonal matrix $ O $ and a tunable $\alpha$; while the matrix $W_h$ is trained without any restriction.

}

\bibliographystyle{IEEEtran}
\bibliography{biblio.bib}

\begin{thebibliography}{10}
\providecommand{\url}[1]{#1}
\csname url@samestyle\endcsname
\providecommand{\newblock}{\relax}
\providecommand{\bibinfo}[2]{#2}
\providecommand{\BIBentrySTDinterwordspacing}{\spaceskip=0pt\relax}
\providecommand{\BIBentryALTinterwordstretchfactor}{4}
\providecommand{\BIBentryALTinterwordspacing}{\spaceskip=\fontdimen2\font plus
\BIBentryALTinterwordstretchfactor\fontdimen3\font minus
  \fontdimen4\font\relax}
\providecommand{\BIBforeignlanguage}[2]{{%
\expandafter\ifx\csname l@#1\endcsname\relax
\typeout{** WARNING: IEEEtran.bst: No hyphenation pattern has been}%
\typeout{** loaded for the language `#1'. Using the pattern for}%
\typeout{** the default language instead.}%
\else
\language=\csname l@#1\endcsname
\fi
#2}}
\providecommand{\BIBdecl}{\relax}
\BIBdecl

\bibitem{lecun2015deep}
Y.~LeCun, Y.~Bengio, and G.~Hinton, ``Deep learning,'' \emph{nature}, vol. 521,
  no. 7553, pp. 436--444, 2015.

\bibitem{rumelhart1986learning}
D.~E. Rumelhart, G.~E. Hinton, and R.~J. Williams, ``Learning representations
  by back-propagating errors,'' \emph{nature}, vol. 323, no. 6088, pp.
  533--536, 1986.

\bibitem{rumelhart1985learning}
D.~{Rumelhart}, G.~{Hinton}, and R.~J. {Williams}, ``Learning internal
  representations by error propagation,'' California Univ San Diego La Jolla
  Inst for Cognitive Science, Tech. Rep., 1985.

\bibitem{ruder2016overview}
S.~Ruder, ``An overview of gradient descent optimization algorithms,''
  \emph{arXiv preprint arXiv:1609.04747}, 2016.

\bibitem{hochreiter1991untersuchungen}
S.~Hochreiter, ``Untersuchungen zu dynamischen neuronalen netzen,''
  \emph{Diploma, Technische Universit{\"a}t M{\"u}nchen}, vol.~91, no.~1, 1991.

\bibitem{bengio1994learning}
Y.~Bengio, P.~Simard, and P.~Frasconi, ``Learning long-term dependencies with
  gradient descent is difficult,'' \emph{IEEE transactions on neural networks},
  vol.~5, no.~2, pp. 157--166, 1994.

\bibitem{glorot2010understanding}
X.~Glorot and Y.~Bengio, ``Understanding the difficulty of training deep
  feedforward neural networks,'' in \emph{Proceedings of the thirteenth
  international conference on artificial intelligence and statistics}.\hskip
  1em plus 0.5em minus 0.4em\relax JMLR Workshop and Conference Proceedings,
  2010, pp. 249--256.

\bibitem{he2015delving}
K.~He, X.~Zhang, S.~Ren, and J.~Sun, ``Delving deep into rectifiers: Surpassing
  human-level performance on imagenet classification,'' in \emph{Proceedings of
  the IEEE international conference on computer vision}, 2015, pp. 1026--1034.

\bibitem{glorot2011deep}
X.~Glorot, A.~Bordes, and Y.~Bengio, ``Deep sparse rectifier neural networks,''
  in \emph{Proceedings of the fourteenth international conference on artificial
  intelligence and statistics}.\hskip 1em plus 0.5em minus 0.4em\relax JMLR
  Workshop and Conference Proceedings, 2011, pp. 315--323.

\bibitem{saxe2013exact}
A.~M. Saxe, J.~L. McClelland, and S.~Ganguli, ``Exact solutions to the
  nonlinear dynamics of learning in deep linear neural networks,'' \emph{arXiv
  preprint arXiv:1312.6120}, 2013.

\bibitem{bertschinger2004real}
N.~Bertschinger and T.~Natschl{\"a}ger, ``Real-time computation at the edge of
  chaos in recurrent neural networks,'' \emph{Neural computation}, vol.~16,
  no.~7, pp. 1413--1436, 2004.

\bibitem{legenstein2007edge}
R.~Legenstein and W.~Maass, ``Edge of chaos and prediction of computational
  performance for neural circuit models,'' \emph{Neural networks}, vol.~20,
  no.~3, pp. 323--334, 2007.

\bibitem{pennington2017resurrecting}
J.~Pennington, S.~S. Schoenholz, and S.~Ganguli, ``Resurrecting the sigmoid in
  deep learning through dynamical isometry: theory and practice,'' \emph{arXiv
  preprint arXiv:1711.04735}, 2017.

\bibitem{mishkin2015all}
D.~Mishkin and J.~Matas, ``All you need is a good init,'' \emph{arXiv preprint
  arXiv:1511.06422}, 2015.

\bibitem{xiao2018dynamical}
L.~Xiao, Y.~Bahri, J.~Sohl-Dickstein, S.~Schoenholz, and J.~Pennington,
  ``Dynamical isometry and a mean field theory of cnns: How to train
  10,000-layer vanilla convolutional neural networks,'' in \emph{International
  Conference on Machine Learning}.\hskip 1em plus 0.5em minus 0.4em\relax PMLR,
  2018, pp. 5393--5402.

\bibitem{pascanu2013difficulty}
R.~Pascanu, T.~Mikolov, and Y.~Bengio, ``On the difficulty of training
  recurrent neural networks,'' in \emph{International conference on machine
  learning}.\hskip 1em plus 0.5em minus 0.4em\relax PMLR, 2013, pp. 1310--1318.

\bibitem{ioffe2015batch}
S.~Ioffe and C.~Szegedy, ``Batch normalization: Accelerating deep network
  training by reducing internal covariate shift,'' in \emph{International
  conference on machine learning}.\hskip 1em plus 0.5em minus 0.4em\relax PMLR,
  2015, pp. 448--456.

\bibitem{srivastava2015highway}
R.~K. Srivastava, K.~Greff, and J.~Schmidhuber, ``Highway networks,''
  \emph{arXiv preprint arXiv:1505.00387}, 2015.

\bibitem{he2016deep}
K.~He, X.~Zhang, S.~Ren, and J.~Sun, ``Deep residual learning for image
  recognition,'' in \emph{Proceedings of the IEEE conference on computer vision
  and pattern recognition}, 2016, pp. 770--778.

\bibitem{veit2016residual}
A.~Veit, M.~J. Wilber, and S.~Belongie, ``Residual networks behave like
  ensembles of relatively shallow networks,'' \emph{Advances in neural
  information processing systems}, vol.~29, 2016.

\bibitem{yang2017mean}
G.~Yang and S.~Schoenholz, ``Mean field residual networks: On the edge of
  chaos,'' \emph{Advances in neural information processing systems}, vol.~30,
  2017.

\bibitem{de2020batch}
S.~De and S.~Smith, ``Batch normalization biases residual blocks towards the
  identity function in deep networks,'' \emph{Advances in Neural Information
  Processing Systems}, vol.~33, pp. 19\,964--19\,975, 2020.

\bibitem{hochreiter1997long}
S.~Hochreiter and J.~Schmidhuber, ``Long short-term memory,'' \emph{Neural
  computation}, vol.~9, no.~8, pp. 1735--1780, 1997.

\bibitem{cho2014properties}
K.~Cho, B.~Van~Merri{\"e}nboer, D.~Bahdanau, and Y.~Bengio, ``On the properties
  of neural machine translation: Encoder-decoder approaches,'' \emph{arXiv
  preprint arXiv:1409.1259}, 2014.

\bibitem{chen2018dynamical}
M.~Chen, J.~Pennington, and S.~Schoenholz, ``Dynamical isometry and a mean
  field theory of rnns: Gating enables signal propagation in recurrent neural
  networks,'' in \emph{International Conference on Machine Learning}.\hskip 1em
  plus 0.5em minus 0.4em\relax PMLR, 2018, pp. 873--882.

\bibitem{gilboa2019dynamical}
D.~Gilboa, B.~Chang, M.~Chen, G.~Yang, S.~S. Schoenholz, E.~H. Chi, and
  J.~Pennington, ``Dynamical isometry and a mean field theory of lstms and
  grus,'' \emph{arXiv preprint arXiv:1901.08987}, 2019.

\bibitem{le2015simple}
Q.~V. Le, N.~Jaitly, and G.~E. Hinton, ``A simple way to initialize recurrent
  networks of rectified linear units,'' \emph{arXiv preprint arXiv:1504.00941},
  2015.

\bibitem{arjovsky2016unitary}
M.~Arjovsky, A.~Shah, and Y.~Bengio, ``Unitary evolution recurrent neural
  networks,'' in \emph{International Conference on Machine Learning}.\hskip 1em
  plus 0.5em minus 0.4em\relax PMLR, 2016, pp. 1120--1128.

\bibitem{wisdom2016full}
S.~Wisdom, T.~Powers, J.~R. Hershey, J.~L. Roux, and L.~Atlas, ``Full-capacity
  unitary recurrent neural networks,'' \emph{arXiv preprint arXiv:1611.00035},
  2016.

\bibitem{lezcano2019cheap}
M.~Lezcano-Casado and D.~Mart{\i}nez-Rubio, ``Cheap orthogonal constraints in
  neural networks: A simple parametrization of the orthogonal and unitary
  group,'' in \emph{International Conference on Machine Learning}.\hskip 1em
  plus 0.5em minus 0.4em\relax PMLR, 2019, pp. 3794--3803.

\bibitem{helfrich2018orthogonal}
K.~Helfrich, D.~Willmott, and Q.~Ye, ``Orthogonal recurrent neural networks
  with scaled cayley transform,'' in \emph{International Conference on Machine
  Learning}.\hskip 1em plus 0.5em minus 0.4em\relax PMLR, 2018, pp. 1969--1978.

\bibitem{lezcano2019trivializations}
M.~Lezcano~Casado, ``Trivializations for gradient-based optimization on
  manifolds,'' \emph{Advances in Neural Information Processing Systems},
  vol.~32, pp. 9157--9168, 2019.

\bibitem{jing2017tunable}
L.~Jing, Y.~Shen, T.~Dubcek, J.~Peurifoy, S.~Skirlo, Y.~LeCun, M.~Tegmark, and
  M.~Solja{\v{c}}i{\'c}, ``Tunable efficient unitary neural networks (eunn) and
  their application to rnns,'' in \emph{International Conference on Machine
  Learning}.\hskip 1em plus 0.5em minus 0.4em\relax PMLR, 2017, pp. 1733--1741.

\bibitem{vorontsov2017orthogonality}
E.~Vorontsov, C.~Trabelsi, S.~Kadoury, and C.~Pal, ``On orthogonality and
  learning recurrent networks with long term dependencies,'' in
  \emph{International Conference on Machine Learning}.\hskip 1em plus 0.5em
  minus 0.4em\relax PMLR, 2017, pp. 3570--3578.

\bibitem{kerg2019non}
G.~Kerg, K.~Goyette, M.~P. Touzel, G.~Gidel, E.~Vorontsov, Y.~Bengio, and
  G.~Lajoie, ``Non-normal recurrent neural network (nnrnn): learning long time
  dependencies while improving expressivity with transient dynamics,''
  \emph{arXiv preprint arXiv:1905.12080}, 2019.

\bibitem{kusupati2019fastgrnn}
A.~Kusupati, M.~Singh, K.~Bhatia, A.~Kumar, P.~Jain, and M.~Varma, ``Fastgrnn:
  A fast, accurate, stable and tiny kilobyte sized gated recurrent neural
  network,'' \emph{arXiv preprint arXiv:1901.02358}, 2019.

\bibitem{li2018independently}
S.~Li, W.~Li, C.~Cook, C.~Zhu, and Y.~Gao, ``Independently recurrent neural
  network (indrnn): Building a longer and deeper rnn,'' in \emph{Proceedings of
  the IEEE conference on computer vision and pattern recognition}, 2018, pp.
  5457--5466.

\bibitem{lukovsevivcius2009reservoir}
M.~Luko{\v{s}}evi{\v{c}}ius and H.~Jaeger, ``Reservoir computing approaches to
  recurrent neural network training,'' \emph{Computer Science Review}, vol.~3,
  no.~3, pp. 127--149, 2009.

\bibitem{lukovsevivcius2012practical}
M.~Luko{\v{s}}evi{\v{c}}ius, ``A practical guide to applying echo state
  networks,'' in \emph{Neural networks: Tricks of the trade}.\hskip 1em plus
  0.5em minus 0.4em\relax Springer, 2012, pp. 659--686.

\bibitem{rotman2020shuffling}
M.~Rotman and L.~Wolf, ``Shuffling recurrent neural networks,'' \emph{arXiv
  preprint arXiv:2007.07324}, 2020.

\bibitem{chang2019antisymmetricrnn}
B.~Chang, M.~Chen, E.~Haber, and E.~H. Chi, ``Antisymmetricrnn: A dynamical
  system view on recurrent neural networks,'' \emph{arXiv preprint
  arXiv:1902.09689}, 2019.

\bibitem{erichson2020lipschitz}
N.~B. Erichson, O.~Azencot, A.~Queiruga, L.~Hodgkinson, and M.~W. Mahoney,
  ``Lipschitz recurrent neural networks,'' \emph{arXiv preprint
  arXiv:2006.12070}, 2020.

\bibitem{rusch2020coupled}
T.~K. Rusch and S.~Mishra, ``Coupled oscillatory recurrent neural network
  (cornn): An accurate and (gradient) stable architecture for learning long
  time dependencies,'' \emph{arXiv preprint arXiv:2010.00951}, 2020.

\bibitem{voelker2019legendre}
A.~R. Voelker, I.~Kaji{\'c}, and C.~Eliasmith, ``Legendre memory units:
  Continuous-time representation in recurrent neural networks,'' 2019.

\bibitem{chandar2019towards}
S.~Chandar, C.~Sankar, E.~Vorontsov, S.~E. Kahou, and Y.~Bengio, ``Towards
  non-saturating recurrent units for modelling long-term dependencies,'' in
  \emph{Proceedings of the AAAI Conference on Artificial Intelligence},
  vol.~33, no.~01, 2019, pp. 3280--3287.

\bibitem{chang2017dilated}
S.~Chang, Y.~Zhang, W.~Han, M.~Yu, X.~Guo, W.~Tan, X.~Cui, M.~Witbrock,
  M.~Hasegawa-Johnson, and T.~S. Huang, ``Dilated recurrent neural networks,''
  \emph{arXiv preprint arXiv:1710.02224}, 2017.

\bibitem{wang2016recurrent}
Y.~Wang and F.~Tian, ``Recurrent residual learning for sequence
  classification,'' in \emph{Proceedings of the 2016 conference on empirical
  methods in natural language processing}, 2016, pp. 938--943.

\bibitem{yue2018residual}
B.~Yue, J.~Fu, and J.~Liang, ``Residual recurrent neural networks for learning
  sequential representations,'' \emph{Information}, vol.~9, no.~3, p.~56, 2018.

\bibitem{mezzadri2006generate}
F.~Mezzadri, ``How to generate random matrices from the classical compact
  groups,'' \emph{arXiv preprint math-ph/0609050}, 2006.

\bibitem{haykin2010neural}
S.~Haykin, \emph{Neural networks and learning machines, 3/E}.\hskip 1em plus
  0.5em minus 0.4em\relax Pearson Education India, 2010.

\bibitem{kingma2014adam}
D.~P. Kingma and J.~Ba, ``Adam: A method for stochastic optimization,''
  \emph{arXiv preprint arXiv:1412.6980}, 2014.

\bibitem{ceni2020echo}
A.~Ceni, P.~Ashwin, L.~Livi, and C.~Postlethwaite, ``The echo index and
  multistability in input-driven recurrent neural networks,'' \emph{Physica D:
  Nonlinear Phenomena}, vol. 412, p. 132609, 2020.

\bibitem{jaeger2007optimization}
H.~Jaeger, M.~Luko{\v{s}}evi{\v{c}}ius, D.~Popovici, and U.~Siewert,
  ``Optimization and applications of echo state networks with leaky-integrator
  neurons,'' \emph{Neural networks}, vol.~20, no.~3, pp. 335--352, 2007.

\bibitem{oaesn2022neuralnet2022}
A.~Ceni and C.~Gallicchio, ``Orthogonal additive echo state networks,''
  \emph{arXiv preprint}, 2022.

\bibitem{ceni2022esann}
A.~{Ceni} and C.~{Gallicchio}, ``Orthogonality in additive reservoir
  computing,'' \emph{arXiv preprint}, 2022.

\bibitem{li2019deep}
S.~Li, W.~Li, C.~Cook, and Y.~Gao, ``Deep independently recurrent neural
  network (indrnn),'' \emph{arXiv preprint arXiv:1910.06251}, 2019.

\bibitem{cooijmans2016recurrent}
T.~Cooijmans, N.~Ballas, C.~Laurent, {\c{C}}.~G{\"u}l{\c{c}}ehre, and
  A.~Courville, ``Recurrent batch normalization,'' \emph{arXiv preprint
  arXiv:1603.09025}, 2016.

\end{thebibliography}













\vspace{11pt}

 
\vspace{11pt}

\vspace{-33pt}
\begin{IEEEbiographynophoto}{Andrea Ceni}
received the M.Sc. degree in Mathematics cum laude from Università degli Studi di Firenze, Italy, in 2017, and the Ph.D. degree from the Department of Computer Science of the University of Exeter, UK, in 2021. He has been a Postdoctoral Research Associate at CEMPS, University of Exeter. Currently, he is a Research Fellow at the Department of Computer Science of the University of Pisa, Italy.
\end{IEEEbiographynophoto}

\vfill

\end{document}